\documentclass[10pt,twocolumn,letterpaper]{article}

\usepackage[pagenumbers]{cvpr}

\usepackage{float}
\usepackage{pifont}
\usepackage{listings}
\usepackage{array}
\usepackage{multirow}
\usepackage{colortbl}

\definecolor{cvprblue}{rgb}{0.21,0.49,0.74}
\definecolor{qualitative_green}{RGB}{0,176,80}
\definecolor{qualitative_orange}{RGB}{255,192,0}
\definecolor{qualitative_blue}{RGB}{68,114,196}
\definecolor{codegreen}{rgb}{0,0.6,0}
\definecolor{negativeorange}{RGB}{214, 39, 40}
\definecolor{LightBlue}{rgb}{0.68, 0.85, 0.9}
\usepackage[pagebackref,breaklinks,colorlinks,allcolors=cvprblue]{hyperref}

\def\paperID{3277}
\def\confName{CVPR}
\def\confYear{2026}

\newcommand{\xmark}{\ding{55}}
\newcommand{\modelname}{Ego2ExoVLM\xspace}
\newcommand{\langdistexpansion}{Ego2Exo Sequence Distillation}
\newcommand{\egotokensexpansion}{Ego Adaptive Visual Tokens}
\newcommand{\benchname}{Ego-in-Exo Perception\xspace}

\title{
    \begin{minipage}{0.125\textwidth}
        \includegraphics[width=\textwidth]{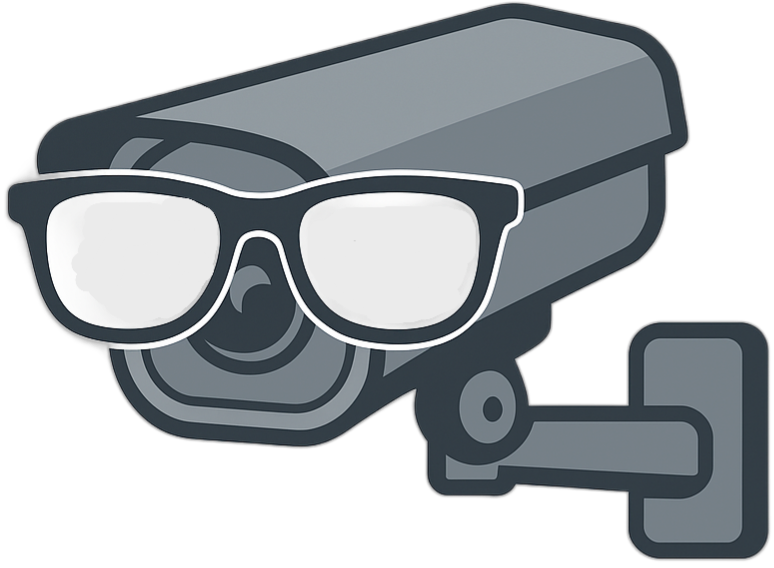}
    \end{minipage}
    \hspace{-0.08\textwidth}
    \begin{minipage}{0.85\textwidth}
        \centering
        From My View to Yours: Learning Egocentric Cues from\\Exocentric Video using Privileged Egocentric Supervision
    \end{minipage}
}

\author{\textbf{Dominick Reilly}$^{1}$ \quad \textbf{Manish Kumar Govind}$^{1}$ \quad \textbf{Le Xue}$^{2}$ \quad \textbf{Srijan Das}$^{1}$ \vspace{0.1cm} \\
$^{1}$University of North Carolina at Charlotte \quad
$^{2}$Elorian AI \vspace{0.1cm} \\
{\tt\small dreilly1@charlotte.edu}\\
\url{https://github.com/dominickrei/EgoExo4ADL}}

\begin{document}
\twocolumn[{
\renewcommand\twocolumn[1][]{#1}%
\maketitle
    \vspace{-10pt}
    \centering
    \scalebox{0.95}{
    \captionsetup{type=figure}
    \includegraphics[width=\textwidth]{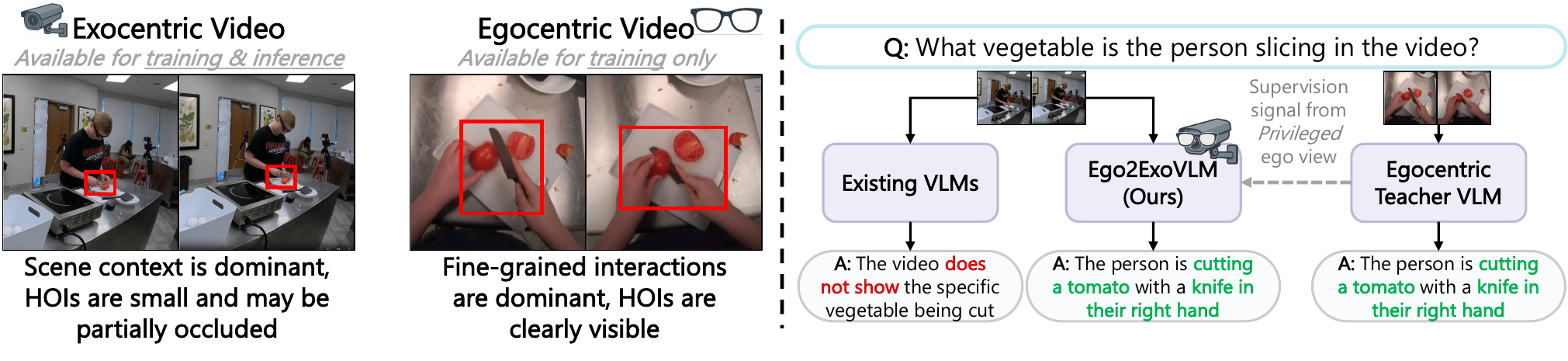}
}
\setcounter{figure}{0} \vspace{-0.1in}
    \captionof{figure}{
    \textbf{Left:} Egocentric and exocentric viewpoints emphasize different visual cues. In exocentric views, the hand-object interactions defining the action appears small or partially occluded, while egocentric views reveal these cues clearly. Red boxes highlight the interaction region.
    \textbf{Right:} Existing VLMs observing exocentric video struggle to capture egocentric properties. \modelname leverages ego-exo video pairs during training, where the egocentric view acts as \textit{privileged} supervision available only during training, enabling VLMs to recover egocentric properties from exocentric video.
    }
    \label{fig:teaser}
    \vspace{0.4cm}
}]

\begin{abstract}
\vspace*{-0.8cm}

\noindent Vision Language Models (VLMs) have achieved strong performance across diverse video understanding tasks. However, their viewpoint invariant training limits their ability to understand egocentric properties (e.g., human-object interactions) from exocentric video observations. This limitation is critical for applications such as Activities of Daily Living (ADL) monitoring, where understanding egocentric properties is essential yet egocentric cameras are impractical to deploy, making it impossible to simply collect egocentric data at test time. To address this challenge, we propose \modelname, a VLM framework that learns to infer egocentric properties from exocentric videos. Our key insight is that time-synchronized ego-exo video pairs can be leveraged during training, where the egocentric viewpoint provides \textit{privileged} supervision (rich egocentric signal available only at training time). \modelname~achieves this through two components: \textbf{\langdistexpansion}, which transfers egocentric reasoning through a language-level sequence distillation objective, and \textbf{\egotokensexpansion}, which encourages the model to surface relevant interaction cues within exocentric video representations. To measure this capability, we introduce \textbf{\benchname}, a benchmark designed to evaluate the understanding of egocentric properties from exocentric videos. \modelname~is evaluated on 10 tasks across \benchname and existing ADL benchmarks, achieving state-of-the-art results on the ADL-X benchmark suite and outperforming strong baselines on our proposed benchmark.
\end{abstract}

\vspace{-0.8cm}
\section{Introduction}
\label{sec:intro}

Vision Language Models (VLMs)~\cite{openai2025thinkwithimages, Qwen2.5-VL, damonlpsg2025videollama3, xue2025blip3} have achieved strong performance across a wide range of multi-modal understanding tasks, from open-ended video question answering~\cite{socratic, mmbench} to complex spatial reasoning~\cite{lai2023lisa,Ranasinghe2024LearningTL}. Existing VLMs for video understanding work by coupling Large Language Models (LLMs)~\cite{qwen2025qwen25technicalreport, meta2024llama3herdmodels} together with pretrained vision encoders~\cite{radford2021openaiclip, zhai2023siglipv1} to enable powerful cross-modal reasoning capabilities. In practice, these models are primarily trained on large-scale video-instruction datasets consisting of both egocentric and exocentric viewpoints~\cite{zhang2024llavavideo, chen2024sharegpt4video, maaz2024videogptplus}. Applying such VLMs to recognize Activities of Daily Living (ADL) from fixed monitoring cameras is a particularly promising direction, with applications in healthcare (e.g., monitoring the elderly, assessing cognitive decline) and assistive robotics.

However, these models struggle in ADL scenarios where the activities of interest are largely defined by \textcolor{codegreen}{egocentric properties}, i.e., \textit{fine-grained scene content} and \textit{human-object interactions} (HOI)~\cite{he2026bridgingperspectives, cvpr2025egoexo4d, sigurdsson2018charades-ego}.
Egocentric and exocentric viewpoints naturally emphasize different visual cues: egocentric views expose fine-grained hand-object interactions and manipulation details, while exocentric views emphasize scene layout, body motion, and contextual information.
Crucially, many egocentric interaction cues (e.g., which object is being manipulated) remain present but visually subtle in the exocentric view, making them difficult for models trained only on exocentric supervision to recover.
As a result, models trained primarily on exocentric observations often fail to capture the egocentric properties that define many ADL.
We identify two primary causes: (i) existing VLMs~\cite{reilly2025llavidal} are trained on instruction-video pairs derived from exocentric ADL datasets~\cite{liu2019ntu, jang2020etri, das2019toyotasmarthome} that emphasize high-level scene context, full-body motion, and environmental details, thereby overlooking egocentric properties that are less visually salient; and (ii) they rarely exploit the corresponding egocentric videos at inference, in which egocentric properties are more directly observable.
This is illustrated in Figure~\ref{fig:teaser}, where existing VLMs observing exo videos fail to capture egocentric properties, while ego-observing VLMs readily capture these details. However, these egocentric observations are difficult to obtain in real-world ADL scenarios, where wearable cameras are impractical to deploy.
This motivates a key question: \textit{how can we train VLMs to infer egocentric properties from exocentric video inputs?}

Consequently, we present \textbf{\modelname}, a novel framework that leverages time-synchronized ego-exo video pairs to transfer egocentric understanding to models observing exocentric videos.
Unlike standard instruction tuning~\cite{liu2023_llava} or distillation~\cite{quattrocchi2024_synch-all-you-need}, where both teacher and student observe the same video, our approach leverages a \textit{privileged} egocentric viewpoint that exposes fine-grained interaction cues which are present in the scene but not visually prominent from the exocentric view.
By learning from this privileged supervision during training, \modelname encourages VLMs operating on exocentric inputs to recover egocentric properties that would otherwise remain latent.
At inference, \modelname takes \textit{only} exocentric video as input and generates outputs that emulate the ego-observing teacher, which naturally captures the egocentric properties we desire to learn from exocentric videos.

\modelname~introduces two complementary components to facilitate ego-to-exo knowledge transfer.
(i) \textbf{\langdistexpansion}: we transfer egocentric reasoning to exocentric observations through language-level distillation, encouraging the model to generate responses consistent with those produced from the privileged egocentric viewpoint. However, response consistency alone is insufficient due to the large ego-exo viewpoint gap, where many interaction cues occupy only a small portion of the exocentric view.
(ii) \textbf{\egotokensexpansion}: to help the model recover these cues we take inspiration from \textit{visual probing} approaches~\cite{reilly2026viscop}, introducing a set of learnable tokens that interact with intermediate visual representations and encourage the extraction of discriminative spatio-temporal regions corresponding to egocentric interactions.
Together, these components enable \textbf{\modelname} to leverage privileged egocentric supervision during training to recover egocentric properties from purely exocentric inputs. Further, to evaluate this capability, we introduce \textbf{\benchname}, a benchmark designed to measure egocentric understanding from exocentric videos. The benchmark leverages time-synchronized ego-exo video pairs where questions are derived from egocentric observations while evaluation is performed using only the paired exocentric videos.

To the best of our knowledge, \modelname is the first VLM that explicitly learns exocentric representations from privileged egocentric supervision.
While the inverse direction, i.e., exo-to-ego transfer, has been explored, works exploring ego-to-exo transfer remain limited, particularly within the VLM domain.
Importantly, the two directions address fundamentally different cues: exo$\rightarrow$ego transfer primarily propagates coarse contextual information such as scene layout and body motion, whereas ego$\rightarrow$exo transfer requires recovering subtle interaction cues such as hand-object interactions that are often visually ambiguous from exocentric viewpoints and therefore difficult to learn through exocentric supervision alone.
Prior approaches for cross-view knowledge transfer are designed for vision-language encoders (e.g., CLIP~\cite{radford2021openaiclip}), and mostly rely on feature-level knowledge transfer in the visual space~\cite{quattrocchi2024_synch-all-you-need, luo2025viewpointrosetta, ohkawa2023_exo2egodvc}. Interestingly, we find that such feature-level knowledge transfer is ineffective in VLMs (Section \ref{sec:experiments_analysis}); instead, we find that \textit{language} serves as a better medium to bridge the knowledge gap between viewpoints.
Some recent works explore cross-view transfer without time-synchronized ego-exo data by relying on pseudo-paired videos~\cite{ohkawa2023_exo2egodvc, xu2024retrievalaugmented} or synthetically generated ego viewpoints~\cite{EMBED}. While such strategies can expand training data, synthetic ego generation is often impractical in ADL settings where activities are captured by fixed monitoring cameras and human-object interactions appear at low spatial resolutions. We therefore focus on the realistic ADL scenario with time-synced ego-exo data; however, we later demonstrate (in Section \ref{sec:unpaired_egoexo}) that \modelname~remains effective even when trained with non-synchronized ego-exo supervision. We summarize our contributions as follows:
\begin{itemize}
    \item We introduce \textbf{\modelname}, the first VLM framework trained with time-synchronized ego-exo video pairs to infer egocentric properties within exocentric inputs, addressing the underexplored problem of ego$\rightarrow$exo transfer for ADL understanding.

    \item We propose two complementary components for enabling cross-view transfer in VLMs: \textbf{\langdistexpansion}, that transfers egocentric reasoning through language-level sequence distillation from a privileged egocentric viewpoint, and \textbf{\egotokensexpansion}, a set of learnable tokens to surface ego-relevant spatio-temporal cues from exocentric videos.

    \item We introduce \textbf{\benchname}, a benchmark of 3.9k MCQs designed to evaluate egocentric understanding from exocentric videos, and validate \modelname~across this benchmark and the ADL-X benchmark suite.
\end{itemize}

\section{Related Work}
\label{sec:related_work}
\noindent\textbf{Vision Language Models for Video}\quad
Advancements in Large Language Models~\cite{llama, gpt, vicuna2023} and large-scale video-text datasets~\cite{li2024_llava-next-interleave, zhang2024llavavideo, zhang2024_llava-hound-dpo, xu2024_slowfast-llava} have led to Vision Language Models (VLMs)~\cite{videollava, videollama, videochatgpt, jin2023chatunivi, li2024_llava-onevision, zhang2024_llava-hound-dpo} with impressive video understanding capabilities.
Most existing VLMs are trained on datasets containing both egocentric and exocentric videos~\cite{ego4d,zhang2024llavavideo}, allowing them to exhibit egocentric understanding only when the input itself is egocentric. However, these models fail to demonstrate egocentric reasoning when observing the same activities from an exocentric viewpoint. Our work is the first to explicitly address this limitation by developing a VLM that transfers egocentric understanding to exocentric videos through paired, time-synchronized ego-exo supervision.

\noindent\textbf{Ego-Exo Video Representation Learning.}\quad
Various prior works have explored video understanding from ego-exo views, and can be categorized~\cite{thatipelli2024_exo2ego-survey} into joint-learning and view transfer approaches.
Joint learning approaches~\cite{xu2018_egoexo-reid, sigurdsson2018_actor-observer, xu2024retrievalaugmented, ohkawa2023_exo2egodvc, yu2019_egoexo-joint-attention, xue2023_egoexo-ae2} aim to learn a unified representation space for both views. For example, Actor and Observer~\cite{sigurdsson2018_actor-observer} trains a dual-stream CNN to contrastively align ego and exo features, while AE2~\cite{xue2023_egoexo-ae2} uses temporal alignment as a contrastive learning objective.
In real-world scenarios where only a single view is available for inference, view transfer approaches~\cite{ardeshir2018_egoexo, li2021_egoexo-transfer, xu2023_egoexo-POV, wang2023_egoexo-SUML, quattrocchi2024_synch-all-you-need, rai2024_egoexo-objectaffordance} aim to leverage knowledge from one view to enhance understanding of the other. For example, Ego-Exo~\cite{li2021_egoexo-transfer} uses egocentric auxiliary tasks to pre-train a 3D-CNN to learn egocentric properties from exocentric videos, Quattrocchi \textit{et al.}~\cite{quattrocchi2024_synch-all-you-need} distills the visual features from an exo-trained teacher to an ego student, EMBED~\cite{EMBED} trains an encoder with synthetic ego data using hand crops derived from exo videos, and ViewpointRosetta~\cite{luo2025viewpointrosetta} uses diffusion models to learn a mapping between the visual feature spaces of ego and exo.
Recent cross-view translation methods~\cite{park2026egoworld, luo2024exo2ego, mahdi2026syn2seq} are also related, but they aim to synthesize or structurally reconstruct observations across viewpoints, whereas our method targets understanding rather than pixel generation.
Most of these approaches focus exclusively on transferring knowledge from the exocentric to egocentric viewpoint, and are not designed for VLMs. In contrast, our work investigates the opposite direction of transferring knowledge from the egocentric to exocentric viewpoint in VLMs.

\section{VLM Preliminaries}
\label{sec:preliminary}
The standard VLM is composed of a vision encoder $E(\cdot)$, a vision-language connector $\phi(\cdot)$, and a language model $LM(\cdot)$.
Given a video $\boldsymbol{V}$, natural language query $\mathbf{Q}$, and a corresponding answer $\mathbf{A}$, the vision encoder first extracts frame-level spatio-temporal tokens from the video
$$\mathbf{X} = E(\boldsymbol{V}) \in \mathbb{R}^{T \times N \times d_v},$$
where $T$ is the number of frames sampled from the video, $N$ is the number of spatial patches, and $d_v$ is the embedding dimension of the vision encoder. These tokens are then mapped to the language model's embedding space, $d_\text{lm}$, using the vision-language connector:
$$\mathbf{Z} = \phi(\mathbf{X}) \in \mathbb{R}^{T \times N \times d_\text{lm}},$$
Conditioned on these projected visual tokens, the training objective of the VLM is to auto-regressively maximize the likelihood of the answer given the video and query
\begin{equation}
\mathcal{L}_\text{VLM} = - \sum_{m=1}^{M} \log P(\mathbf{a}_m \,|\, \mathbf{A}_{<m}, \mathbf{Q}, \mathbf{Z}),
\end{equation}
where $M$ is the number of tokens in the answer, and $\mathbf{A}_{<m} = (\mathbf{a}_1, \ldots, \mathbf{a}_{m-1})$ denotes the sequence of tokens prior to the auto-regressive decoding step $m$.

\section{Task Formulation}

\label{sec:task_formulation}
Given time-synchronized pairs of ego-exo activity videos $(\boldsymbol{V}_\text{ego}, \boldsymbol{V}_\text{exo})$ along with a natural language query $\mathbf{Q}$, we aim to learn a VLM that can accurately respond to the query using only the exocentric video at inference time. Note that the answer $\mathbf{A}$ is not explicitly given at training time.
Formally, during training, we have access to triplets $(\boldsymbol{V}_\text{ego}, \boldsymbol{V}_\text{exo}, \mathbf{Q})$, but at test time, the model only has access to $(\boldsymbol{V}_\text{exo}, \mathbf{Q})$.
The challenge of this task lies in the fact that the language queries target egocentric properties of the video, which are not salient from the exocentric viewpoint.
The necessity of this task is grounded in the understanding of Activities of Daily Living, where such egocentric properties define the activities of interest; however, the egocentric view is rarely available in real-world ADL scenarios.
To solve this, we leverage time-synchronized ego-exo video pairs during training to transfer egocentric understanding from an ego-observing teacher VLM to an exo-observing student VLM.

\section{Proposed Method}
\label{sec:methods}

\begin{figure*}[t!]
    \centering
    \begin{minipage}[t]{0.48\linewidth}
        \centering
        \includegraphics[width=\linewidth]{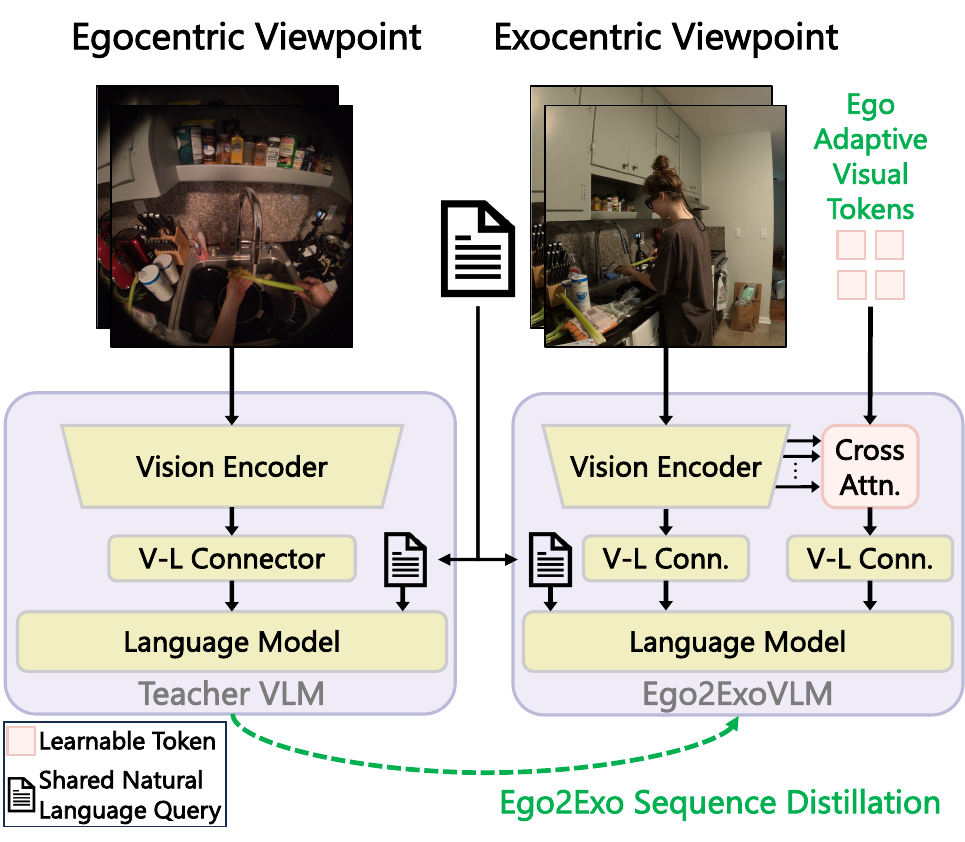}
        \caption{\textbf{Overview of our proposed \modelname.}
        A teacher VLM observing egocentric viewpoint videos transfers knowledge to a student VLM through \langdistexpansion.
        The student VLM is augmented with \egotokensexpansion, which are cross-attended with exocentric visual features.}
        \label{fig:method}
    \end{minipage}
    \hfill
    \begin{minipage}[t]{0.48\linewidth}
        \centering
        \includegraphics[width=\linewidth]{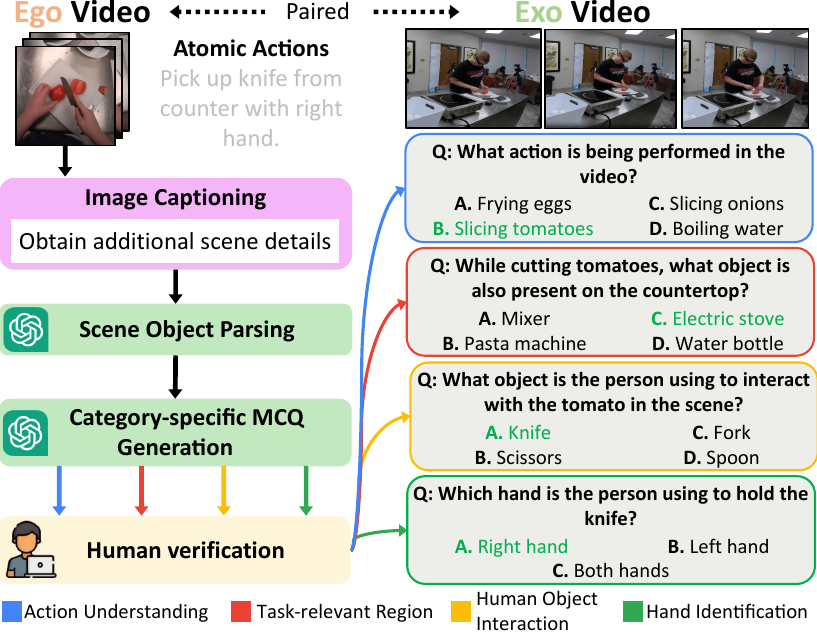}
        \caption{\textbf{\benchname data curation pipeline.}
        The egocentric viewpoint is used to generate question-answer pairs for the corresponding time-synched exocentric viewpoint. In other words, at inference we query the exocentric video with egocentric questions. Additional examples available in the Appendix.}
        \label{fig:egoinexo_diagram}
    \end{minipage}
    \vspace{-0.25cm}
\end{figure*}

This section introduces our proposed method, \textbf{\modelname}, which addresses ego-to-exo viewpoint transfer through knowledge distillation. \modelname~training follows a teacher-student paradigm (see Figure~\ref{fig:method}) and comprises two complementary components:
(1) \textbf{\langdistexpansion} that transfers egocentric semantics from an egocentric teacher to an exocentric student, and
(2) \textbf{\egotokensexpansion} that enhances the student's ability to acquire egocentric semantics from the teacher.

Both the teacher and student are VLMs sharing the architecture described in Section \ref{sec:preliminary}.
The \textbf{teacher VLM} $\mathcal{T}$ processes the egocentric video to generate an answer that implicitly captures egocentric properties, which then serves as a source of egocentric supervision for the exocentric \textbf{student VLM} $\mathcal{S}$, which processes the corresponding exocentric video. During training, the teacher remains frozen and serves as a high-level semantic bridge between the two viewpoints.

\subsection{\langdistexpansion}
\label{sec:method_viewtransfer}
Our goal is to leverage the time-synchronized ego-exo videos $(\boldsymbol{V}_\text{ego}, \boldsymbol{V}_\text{exo})$ to transfer egocentric understanding from $\mathcal{T}$ to $\mathcal{S}$.
In \langdistexpansion, the teacher observes the activity from the egocentric viewpoint to generate a language sequence capturing the egocentric properties of the activity, while the student observes the activity from the exocentric viewpoint and is supervised by the teacher to reproduce its outputs.
Formally, the teacher VLM processes the egocentric video and query to generate the language sequence $\mathbf{A}_{ego} = \mathcal{T}(\boldsymbol{V}_{ego}, \mathbf{Q})$, which serves as a pseudo-ground truth answer to guide the student VLM. During training, the objective of the student is to auto-regressively maximize the likelihood of $\mathbf{A}_{ego}$
\begin{equation}
\mathcal{L}_{\text{VLM}} = - \sum_{m=1}^{M} \log P(\mathbf{a}^{ego}_m \,|\, \mathbf{A}^{ego}_{<m}, \mathbf{Q}, \mathbf{Z}^{exo})
\end{equation}
This objective enables cross-view knowledge transfer by enforcing consistency between the egocentric teacher, which implicitly understands the egocentric properties of the video, and the exocentric student.

\subsection{\egotokensexpansion}
\label{sec:method_adaptivetokens}
While the sequence-level distillation described above provides the supervisory signal necessary to transfer viewpoint knowledge from teacher to student, we observe that exocentric videos contain many visual distractors (e.g., background content) that dilute the spatial cues relevant for egocentric understanding, limiting the effectiveness of cross-view transfer. To enhance the effectiveness of this transfer, we introduce a set of learnable \egotokensexpansion~that guide the student VLM to focus on spatial regions carrying ego-relevant cues. These additional tokens enable the student to selectively extract exocentric visual features that better align with the egocentric language sequence. In essence, the ego adaptive tokens are supervised through the egocentric signal to capture spatio-temporal patterns that are less prominent in the exocentric view.

Concretely, the learnable tokens are implemented as \textit{visual probes}~\cite{reilly2026viscop} that are cross-attended with the intermediate representations of the vision encoder, allowing them to leverage both low- and high-level pixel information to facilitate the extraction of discriminative exocentric representations for bridging the gap between the two viewpoints. Letting the encoder output features at layer $\ell$ be $\mathbf{X}^\ell \in \mathbb{R}^{T \times N \times d_v}$ and the learnable ego adaptive tokens be $\mathbf{E}^\ell \in \mathbb{R}^{K \times d_v}$, cross-attention is applied as follows
$$
\mathbf{\tilde{E}}^{\ell} =
\mathrm{softmax}\!\left(
\frac{(\mathbf{E}^{\ell} \boldsymbol{W}_q^{\ell})(\mathbf{X}^{\ell} \boldsymbol{W}_k^{\ell})^{\top}}{\sqrt{d_v}}
\right)
(\mathbf{X}^{\ell} \boldsymbol{W}_v^{\ell}),
$$
where $(\boldsymbol{W}_q^\ell, \boldsymbol{W}_k^\ell, \boldsymbol{W}_v^\ell)$ are layer-specific projection matrices. The tokens after the final layer (denoted $\mathbf{\tilde{E}}$ for simplicity) are projected into the language model's embedding space through a dedicated connector $\phi_{\text{ego}}(\cdot)$ and concatenated with the exocentric visual embeddings before being fed to the language model. Thus, the training objective of \modelname is defined as
\begin{equation}
\mathcal{L}_{\text{VLM}} = - \sum_{m=1}^{M} \log P(\mathbf{a}^{ego}_m \,|\, \mathbf{A}^{ego}_{<m}, \mathbf{Q}, \mathbf{Z}^{exo}, \phi_{\text{ego}}(\mathbf{\tilde{E}}))
\end{equation}
With this objective, the teacher is able to transfer egocentric cues to the student by leveraging the \textit{shared content} of time-synchronized ego-exo videos.

\section{\benchname~Benchmark}
\label{sec:egoperceptionmcq}

In this section, we introduce \benchname, a benchmark containing a total of 3,881 multiple choice questions (MCQs) designed to evaluate egocentric understanding from exocentric video inputs. It consists of time-synchronized videos of skilled human activities captured simultaneously from both egocentric and exocentric viewpoints.
We propose this benchmark to answer the single question: \textit{How well do VLMs capture egocentric properties from exocentric videos?}

The key novelty of \benchname lies in how it is designed to answer this question. We posit that to measure egocentric understanding from exocentric videos, the question-answer (QA) pairs used in our benchmark should be curated using only the corresponding egocentric view. Our intuition behind this is that if QAs are curated based on what is salient in ego, then answering these questions from the exo video alone directly measures the model's ability to capture ego properties from exo videos. Samples from our benchmark are shown in Figure \ref{fig:egoinexo_diagram} (right), with more samples presented in the supplementary materials.

\vspace{-0.15cm}
\subsection{Data Curation}

We curate \benchname from the EgoExo4D~\cite{cvpr2025egoexo4d} dataset, a large-scale dataset of long videos of skilled human activities captured from ego and exo viewpoints.
We only consider videos from the keystep recognition \textit{test} subset of EgoExo4D, and only keep videos depicting ``\textit{cooking}'' activities, as these activities contain a high density and diversity of human-object interactions~\cite{Damen2018EPICKITCHENS}, making them an ideal testbed for measuring the egocentric properties we are interested in. This filtered subset provides us with 2,319 unique videos along with 10K human annotations detailing the atomic actions within each video. Our benchmark contains four tasks that evaluate different aspects of egocentric understanding: Action Understanding, Task-relevant Region, Human-Object Interactions, and Hand Identification.

The videos and annotations are then processed through a multi-stage pipeline, as illustrated in Figure \ref{fig:egoinexo_diagram}, to obtain MCQs. We first extract a list of the objects visible in the egocentric video by applying an image captioning model~\cite{hong2023cogagent} at 5fps and parsing its output using an LLM~\cite{openai2024gpt4ocard} to obtain a list of scene objects. Next, we generate MCQs across the four tasks using an LLM~\cite{openai2024gpt4ocard} with category-specific prompts applied to the atomic action annotations, supplemented with the list of scene objects. We find that supplementing the annotations with this list enables the generation of more diverse question types with challenging negative distractors. Finally, we perform a comprehensive human verification with four participants to filter out hallucinated questions, trivial distractors, and questions that are unanswerable from the exocentric view due to occlusions or poor camera framing, yielding our final benchmark of 3,881 questions. A more detailed description of each step in this pipeline is provided in the Appendix.

\begin{table}[h]
    \centering
    \caption{\textbf{Differentiating from existing benchmarks.} Columns indicate benchmark properties: \textbf{Ego-Exo}: contains paired ego-exo videos, \textbf{Time-sync}: videos are time-synchronized, \textbf{Video-QA}: suitable for evaluating VLMs, \textbf{Ego-curated QA}: video-QA pairs are curated only from the egocentric viewpoint.}
    \vspace{-0.2cm}
    \resizebox{0.98\linewidth}{!}{
    \begin{tabular}{c|cccc}
        \hline
        \textbf{Benchmark} & \textbf{Ego-Exo?} & \textbf{Time-sync?} & \textbf{Video-QA?} & \textbf{Ego-curated QA?} \\
        \hline
        Kinetics~\cite{kinetics600} & \xmark & \xmark & \xmark & \xmark \\
        VideoMME~\cite{fu2025videomme} & \xmark & \xmark & \checkmark & \xmark \\
        \hline
        EgoMCQ~\cite{kevin2022egovlp} & \xmark & \xmark & \xmark & \checkmark \\
        EgoSchema~\cite{egoschema} & \xmark & \xmark & \checkmark & \checkmark \\
        EgoMemoria~\cite{ye2025mmego} & \xmark & \xmark & \checkmark & \checkmark \\
        \hline
        LEMMA~\cite{lemma} & \checkmark & \checkmark & \xmark & \xmark \\
        EgoExoLearn~\cite{huang2024egoexolearn} & \checkmark & \checkmark & \xmark & \xmark \\
        EgoExoBench~\cite{he2025egoexobench} & \checkmark & \checkmark & \checkmark & \xmark \\
        \hline
        \rowcolor{LightBlue}
        \textbf{Ego-in-Exo Perception} & \checkmark & \checkmark & \checkmark & \checkmark \\
        \hline
    \end{tabular}}
    \label{tab:egoinexo_existing}
\end{table}

\vspace{-0.15cm}
\subsection{Differences from Existing Benchmarks}
Table~\ref{tab:egoinexo_existing} compares \benchname with existing video understanding benchmarks. While several benchmarks exist to evaluate exocentric~\cite{kinetics600, fu2025videomme} and egocentric~\cite{kevin2022egovlp, lemma, egoschema, ye2025mmego} understanding, they lack the time-synchronized ego-exo videos. Existing ego-exo benchmarks like LEMMA~\cite{lemma} and EgoExoLearn~\cite{huang2024egoexolearn} provide time-synchronized ego-exo videos, but are not designed to evaluate VLMs. EgoExoBench~\cite{he2025egoexobench} contains time-synchronized ego-exo videos and is designed for VLM evaluation, but uses both viewpoints in its data curation, making it unclear whether VLMs succeed by capturing egocentric properties or by exploiting information already salient in the exocentric view. In contrast, our benchmark uniquely leverages time-synchronized ego-exo videos, curating QA pairs from the egocentric viewpoint alone.

\begin{table}[h]
    \renewcommand{\arraystretch}{1.2}
    \centering
    \caption{\textbf{Bias analysis on \benchname.}}
    \vspace{-0.2cm}
    \resizebox{\linewidth}{!}{
        \begin{tabular}{ccccccc}
        \hline
        \multirow{2}{*}{\textbf{Method}} & \multirow{2}{*}{\textbf{Inputs}} &
            \multirow{2}{*}{\shortstack{\textbf{Action}\\\textbf{Und.}}} &
            \multirow{2}{*}{\shortstack{\textbf{Task}\\\textbf{Regions}}} &
            \multirow{2}{*}{\textbf{HOI}} &
            \multirow{2}{*}{\shortstack{\textbf{Hand}\\\textbf{Ident.}}} &
            \multirow{2}{*}{\textbf{Avg.}} \\
        & & & & & & \\
        \hline
        \multirow{3}{*}{GPT-4.1-Mini}
        & Text only & 56.4 & 56.3 & 55.3 & 54.4 & 55.7 \\
        & Text + Image & 71.4 & 79.8 & 70.3 & 55.7 & 69.3 \\
        & Text + Video & \textbf{77.5} & 76.6 & 75.9 & 59.5 & 72.4 \\
        \midrule
        GPT-5.1 & Text + Video & 74.6 & \textbf{84.8} & \textbf{77.2} & 54.8 & 72.9 \\
        \midrule
        Ego2ExoVLM  & Text + Video & 73.4 & 82.0  & 76.2 & \textbf{65.3} & \textbf{74.2} \\
        \hline
        \end{tabular}}
    \label{tab:benchmark_bias_analysis}
    \vspace{-0.2cm}
\end{table} 

\subsection{Evaluating Visual and Linguistic Bias}

To assess whether Ego-in-Exo Perception can be solved via linguistic priors or simple visual shortcut biases, we evaluate GPT-4.1-Mini and GPT-5.1 under three input settings: \textit{Text only} (question without video), \textit{Text + Image} (single center frame), and \textit{Text + Video} (full video).
As shown in Table~\ref{tab:benchmark_bias_analysis}, performance drops substantially in the Text-only setting (55.7\% Avg.), indicating limited linguistic bias.
Providing only the center frame improves results (69.8\%), but still underperforms full video input (73.4\%), suggesting that temporal information is necessary. Notably, \modelname~achieves higher overall accuracy than GPT-4.1-Mini in the full Video setting (74.2\% vs 73.4\%), demonstrating that the benchmark rewards viewpoint-specific reasoning rather than scale alone and that the benchmark exhibits relatively weak biases, a finding in line with existing video understanding benchmarks~\cite{zohar2025apollo}.

\begin{table}[h]
    \centering
    \caption{\textbf{State-of-the-art comparison on the \benchname benchmark.} Action Und.: \textit{Action Understanding}, Task Regions: \textit{Task Relevant Regions}, HOI: \textit{Human Object Interactions}, Hand Ident.: \textit{Hand Identification}.}
    \vspace{-0.2cm}
    \resizebox{0.98\linewidth}{!}{
    \begin{tabular}{c|ccccc}
        \hline
        \multirow{2}{*}{\textbf{Method}} &
        \multirow{2}{*}{\shortstack{\textbf{Action}\\\textbf{Und.}}} &
        \multirow{2}{*}{\shortstack{\textbf{Task}\\\textbf{Regions}}} &
        \multirow{2}{*}{\textbf{HOI}} &
        \multirow{2}{*}{\shortstack{\textbf{Hand}\\\textbf{Ident.}}} &
        \multirow{2}{*}{\textbf{Avg.}} \\
        & & & & & \\
        \hline
        \rowcolor{gray!20}\multicolumn{6}{c}{\textit{Evaluate on ego viewpoint (upper bound)}} \\
        Teacher ($\mathcal{T}$) & 81.8 & 83.9 & 78.9 & 67.1 & 77.9 \\
        \hline
        \rowcolor{gray!20}\multicolumn{6}{c}{\textit{Evaluate on exo viewpoint}} \\
        LLAVIDAL~\cite{reilly2025llavidal} & 15.2 & 33.3 & 31.0 & 53.8 & 33.3 \\
        VideoLLaMA2~\cite{damonlpsg2024videollama2} & 35.3 & 42.8 & 34.6 & 48.4 & 40.3 \\
        Qwen2-VL~\cite{wang2024qwen2vl} & 38.5 & 48.9 & 37.4 & 56.4 & 45.3 \\
        Qwen2.5-VL~\cite{qwen2025qwen25technicalreport} & 68.2 & 75.9 & 70.8 & 64.8 & 69.9 \\
        VideoLLaMA3~\cite{damonlpsg2025videollama3} & 66.5 & 79.6 & 74.5 & 64.1 & 71.2 \\
        \hline
        EE4D-VideoLLaMA3 & 59.5 & 66.6 & 71.6 & 65.0 & 65.7 \\
        \rowcolor{LightBlue}
        \textbf{\modelname~(Ours)} & \textbf{73.4} & \textbf{82.0} & \textbf{76.2} & \textbf{65.3} & \textbf{74.2} \\
        \hline
    \end{tabular}}
    \label{tab:sota_egoinexo}
    \vspace{-0.2cm}
\end{table}

\section{Experiments}
\label{sec:experiments}
In Section \ref{sec:experiments_implementation}, we detail the experimental setup. We then compare \modelname\ to the state-of-the-art on two activity understanding datasets in Section \ref{sec:experiments_results}. Finally, in Section \ref{sec:experiments_analysis}, we perform ablations on the \benchname benchmark and analyze the learned representations of \modelname.

\begin{table*}[t!]
    \centering
    \caption{\textbf{State-of-the-art comparison on the ADL-X benchmark.} Methods are categorized by image captioning models paired with LLMs and vision language models. Legend for ADL Video Description tasks: [Cor: \textit{Correctness of Information}, Do: \textit{Detail Orientation}, Ctu: \textit{Contextual Understanding}, Tu: \textit{Temporal Understanding}, Con: \textit{Consistency}]}
    \resizebox{0.98\linewidth}{!}{
    \begin{tabular}{c|ccccc|cccccc|cccccc}
         \hline
         \multirow{2}{*}{\textbf{Method}} & \multicolumn{5}{c|}{\textbf{ADL MCQ}} & \multicolumn{6}{c|}{\textbf{ADL-VD (Charades)}} & \multicolumn{6}{c}{\textbf{ADL-VD (Toyota Smarthome)}}\\

        & \textbf{Charades AR} & \textbf{SH AR} & \textbf{LEMMA TC} & \textbf{TSU TC} & \textbf{Avg} & \textbf{Cor} & \textbf{Do} & \textbf{Ctu} & \textbf{Tu} & \textbf{Con} & \textbf{Avg} & \textbf{Cor} & \textbf{Do} & \textbf{Ctu} & \textbf{Tu} & \textbf{Con} & \textbf{Avg} \\
         \hline
         \rowcolor{gray!20}\multicolumn{18}{c}{\textit{Image captioners + LLM}} \\
         CogVLM~\cite{cogvlm} + GPT~\cite{gpt} & 52.3 & 42.5 & 32.0 & 23.6 & 37.6 & 42.0 & 62.0 & 49.6 & 36.5 & 32.8 & 44.6 & 55.2 & 72.0 & 60.6 & 30.2 & 48.5 & 53.3 \\
         CogVLM~\cite{cogvlm} + Llama~\cite{llama} & 52.8 & 43.2 & 32.5 & 22.5 & 37.8 & 40.2 & 61.8 & 49.5 & 36.5 & 33.5 & 44.3 & 49.8 & 66 & 56.6 & 29.8 & 40.2 & 48.5 \\
         BLIP2~\cite{blip2} + GPT~\cite{gpt} & 50.2 & 39.6 & 28.9 & 20.2 & 34.7 & 39.8 & 60.2 & 47.8 & 36.0 & 37.2 & 44.2 & 48.8 & 66.6 & 63.6 & 45.6 & 39.8 & 52.9 \\
         \hline
         \rowcolor{gray!20}\multicolumn{18}{c}{\textit{Vision Language Models}} \\
         VideoChatGPT~\cite{videochatgpt} & 51.0 & 39.6 & 31.4 & 20.9 & 35.7 & 26.1 & 45.2 & 35.6 & 21.4 & 31.2 & 31.9 & 31.2 & 52.8 & 78.2 & 64.8 & 45.6 & 54.5 \\
         VideoLLaMA~\cite{videollama} & 40.2 & 44.8 & 32.6 & 24.6 & 35.6 & 22.2 & 42.5 & 33.8 & 20.2 & 34.5 & 30.6 & 57.8 & 62.0 & 62.4 & 48.2 & 44.4 & 54.9\\
         VideoLLaVA~\cite{videollava} & 41.8 & 49.2 & 30.0 & 25.5 & 36.6 & 23.6 & 46.4 & 34 & 20.6 & 33.5 & 31.6 & 30.8 & 54.8 & 42.4 & 30.4 & 44.5 & 40.6\\
         Chat-UniVi~\cite{jin2023chatunivi} & 53.1 & 48.1 & 32.3 & 36.4 & 42.5 & 36.5 & 54.5 & 46.6 & 32.2 & 35.9 & 41.1 & 56.8 & 66.9 & 79.0 & 50.0 & 56.6 & 61.9 \\
         ADL-X-ChatGPT~\cite{reilly2025llavidal} & 51.0 & 44.5 & 28.6 & 29.5 & 38.4 & 40.6 & 50.6 & 49.8 & 30.6 & 40.2 & 42.4 & 62.4 & 79.4 & 70.8 & 51.2 & 60.4 & 64.8\\
         LLAVIDAL~\cite{reilly2025llavidal} & 55.2 & 48.1 & 34.3 & 38.2 & 44.0 & 45.8 & 64.2 & 57.0 & 36.4 & 39.4 & 48.6 & 66.0 & 86.2 & 79.6 & 50.0 & 72.4 & 70.8 \\
         VideoLLaMA3~\cite{damonlpsg2025videollama3} & 92.0 & 70.6 & 66.6 & 78.3 & 76.9 & 73.5 & 73.7 & 75.8 & 68.6 & 61.6 & 70.6 & 88.1 & 91.8 & 92.6 & 79.2 & \textbf{84.4} & 87.2 \\
         \hline
         EE4D-VideoLLaMA3 & 90.9 & 71.3 & 60.2 & 77.7 & 75.0 & 80.4 & 78.8 & 82.0 & 73.8 & 61.9 & 75.4 & \textbf{91.5} & 92.5 & 95.8 & 79.0 & 73.5 & 86.4 \\
         \rowcolor{LightBlue}
         \textbf{\modelname (Ours)} & \textbf{93.1} & \textbf{72.0} & \textbf{67.4} & \textbf{82.6} & \textbf{78.8} & \textbf{80.5} & \textbf{78.8} & \textbf{82.3} & \textbf{73.8} & \textbf{62.4} & \textbf{75.5} & 90.8 & \textbf{93.1} & \textbf{95.9} & \textbf{80.8} & 82.9 & \textbf{88.7} \\
         \hline
    \end{tabular}}
    \label{tab:sota_adlx}
    \vspace{-0.2cm}
\end{table*}

\subsection{Experimental Setting}
\label{sec:experiments_implementation}
\noindent\textbf{Teacher and Student VLMs.}\quad
Both the teacher and student adopt identical VLM architectures consisting of a SigLIP~\cite{zhai2023siglipv1} vision encoder (embedding dim. $d_v = 1152$) and a Qwen2.5~\cite{qwen2025qwen25technicalreport} language model (embedding dim. $d_\text{lm} = 3584$), initialized from the pretrained weights of VideoLLaMA3~\cite{damonlpsg2025videollama3}. During training, the teacher VLM remains frozen and the following components of the student VLM remain trainable: vision-language connector, cross-attention projection matrices (for ego-adaptive tokens), and low-rank adaptation (LoRA with rank $r = 64$) weights in the LLM are updated. The vision encoder remains frozen in all experiments. Training is conducted on 4 NVIDIA H200 GPUs for 3 epochs, with a learning rate of $1\times10^{-5}$ for the LoRA parameters and $2\times10^{-6}$ for all other parameters.

\noindent\textbf{Student Training Data.}\quad
The student VLM is trained on a total of 9k ego-exo videos from the keystep recognition training set of EgoExo4D~\cite{cvpr2025egoexo4d}. For each ego-exo video pair, we generate three distinct natural language queries (i.e., $\mathbf{Q}$) by first sampling one query from each of three predefined categories: descriptive, temporal, and spatial (provided in supplementary materials).
We then process the sampled query along with the egocentric video with the frozen pre-trained teacher VLM which generates an answer ($\mathbf{A}$). This query-answer pair then serves as supervision to the student VLM, yielding a total of 28k query-answer pairs.

\noindent\textbf{Benchmarks.}\quad
We evaluate \modelname~across 10 tasks that rely on egocentric understanding from exocentric videos. The \textbf{\benchname} benchmark, introduced in this work, is derived from the Ego-Exo4D~\cite{cvpr2025egoexo4d} dataset and consists of 4 MCQ tasks introduced in Section \ref{sec:egoperceptionmcq}: Action Understanding, Task-relevant Regions, Human Object Interactions, and Hand Identification. The \textbf{ADL-X} benchmark~\cite{reilly2025llavidal} evaluates VLM's understanding on multiple Activities of Daily Living (ADL) datasets~\cite{liu2019ntu, das2019toyotasmarthome, lemma, dai2022toyotasmarthomeuntrimmed}. Although ADL videos are recorded only from the exocentric viewpoint, the activities center around hand-object interactions and fine-grained motion details, making ADL a natural setting to assess egocentric understanding from exocentric videos. The benchmark contains 4 MCQ tasks (ADL-MCQ) and 2 video description tasks (ADL-VD). ADL-MCQ contains both Action Recognition (AR) and Temporal Completion (TC) tasks. Accuracy is reported for all MCQ tasks and Video-ChatGPT description metrics~\cite{videochatgpt} are reported for description tasks.

\subsection{Comparison with SoTA}
\label{sec:experiments_results}
\vspace{-0.2cm}
\noindent\textbf{\benchname~Benchmark.}\quad
Table~\ref{tab:sota_egoinexo} compares \modelname~with existing VLMs on our proposed \benchname benchmark. We include the teacher model's performance on egocentric videos as an upper bound (77.9\% average accuracy), representing the level of understanding achievable with direct access to the egocentric viewpoint. We also present a variant of VideoLLaMA3 finetuned on the exocentric videos of Ego-Exo4D (EE4D-VideoLLaMA3). LLAVIDAL~\cite{reilly2025llavidal} struggles significantly (33.3\%), particularly on Action Understanding (15.2\%) despite being trained on ADL data, highlighting that standard exocentric training does not capture the egocentric properties targeted by our benchmark. VideoLLaMA3~\cite{damonlpsg2025videollama3}, whose weights are used to initialize our student VLM, achieves higher accuracy than other baselines, likely due to its large-scale pretraining data. However, \modelname~outperforms this strong baseline by +3.0 \% (71.2\% vs 74.2\%). Comparing \modelname with EE4D-VideoLLaMA3, our method achieves superior performance (74.2\% vs 65.7\%), demonstrating that leveraging time-synchronized ego-exo pairs for cross-view knowledge transfer is more effective than training solely on exo videos from the same dataset.

\noindent\textbf{ADL-X.}\quad
Table~\ref{tab:sota_adlx} compares \modelname~against state-of-the-art methods on the ADL-X benchmark. Image captioning models such as CogVLM~\cite{cogvlm} achieve moderate performance on ADL tasks even when paired with large language models like GPT-4 (44.6\% on Charades Description compared to 75.5\% on \modelname). Among existing VLMs, pre-trained VideoLLaMA3 performs better on ADL tasks, and EE4D-VideoLLaMA3 can further improve performance by leveraging in-domain data. However, our approach surpasses both, reaching 78.8\% on ADL-MCQ, 75.5\% on Charades Descriptions, and 88.7\% on Toyota Smarthome Descriptions. \modelname~also outperforms LLAVIDAL~\cite{reilly2025llavidal}, a VLM specifically tailored for ADL understanding, demonstrating \modelname's understanding of the egocentric properties relevant in ADL.

\begin{table}[h!]
    \centering
    \setlength{\tabcolsep}{5pt}
    \caption{\textbf{Alternative Viewpoint Transfer Strategies.}}
    \vspace{-0.2cm}
    \resizebox{0.98\linewidth}{!}{
    \begin{tabular}{c|ccccc}
        \hline
        \multirow{2}{*}{\textbf{Transfer Strategy}} &
        \multirow{2}{*}{\shortstack{\textbf{Action}\\\textbf{Und.}}} &
        \multirow{2}{*}{\shortstack{\textbf{Task}\\\textbf{Regions}}} &
        \multirow{2}{*}{\textbf{HOI}} &
        \multirow{2}{*}{\shortstack{\textbf{Hand}\\\textbf{Ident.}}} &
        \multirow{2}{*}{\textbf{Avg.}} \\
        & & & & & \\
        \hline
        Visual Feature Dist.~\cite{quattrocchi2024_synch-all-you-need} & 61.2 & 77.7 & 68.0 & 64.2 & 67.8 \\
        LLAVIDAL-style~\cite{reilly2025llavidal} & 60.2 & 66.1 & 70.3 & 65.0 & 65.4 \\
        Exo2EgoVLM~\cite{zhang2025exo2egovlm} & 65.9 & 75.1 & 63.8 & 81.1 & 72.5 \\
        EgoAda Tokens (Vis.) & 69.5 & 79.4 & 74.1 & 65.3 & 72.1 \\
        EgoAda Tokens (Lang.) & 69.2 & 78.2 & 72.0 & 64.9 & 71.1 \\
        \hline
        \rowcolor{LightBlue}
        \textbf{\modelname} & \textbf{73.4} & \textbf{82.0} & \textbf{76.2} & \textbf{65.3} & \textbf{74.2} \\
        \hline
    \end{tabular}}
    \label{tab:ablation_diststrat}
    \vspace{-0.2cm}
\end{table}

\begin{table*}[t!]
\renewcommand{\arraystretch}{1.2}
\centering
\begin{minipage}[t]{0.51\linewidth}
    \centering
    \setlength{\tabcolsep}{5pt}
    \caption{\textbf{Ablating \modelname components.} Seq Dist. = \textit{\langdistexpansion}, Ego Tok. = \textit{\egotokensexpansion}.}
    \vspace{-0.2cm}
    \resizebox{0.98\linewidth}{!}{
    \begin{tabular}{c|cc|ccccc}
        \hline
        \multirow{2}{*}{\shortstack{\textbf{Training}\\\textbf{View}}} &
        \multirow{2}{*}{\shortstack{\textbf{Seq}\\\textbf{Dist.}}} &
        \multirow{2}{*}{\shortstack{\textbf{Ego}\\\textbf{Tok.}}} &
        \multirow{2}{*}{\shortstack{\textbf{Action}\\\textbf{Und.}}} &
        \multirow{2}{*}{\shortstack{\textbf{Task}\\\textbf{Regions}}} &
        \multirow{2}{*}{\textbf{HOI}} &
        \multirow{2}{*}{\shortstack{\textbf{Hand}\\\textbf{Ident.}}} &
        \multirow{2}{*}{\textbf{Avg.}} \\
        & & & & & & & \\
        \hline
        Exo & \xmark & \xmark & 59.5 & 66.6 & 71.6 & 65.0 & 65.7 \\
        Ego+Exo & \xmark & \xmark & 59.2 & 68.5 & 71.1 & 65.0 & 66.0 \\
        Exo & \checkmark & \xmark & 67.2 & 67.9 & 73.8 & 65.3 & 68.5 \\
        Exo & \xmark & \checkmark & 70.0 & 79.4 & 74.3 & 65.1 & 72.2 \\
        \hline
        \rowcolor{LightBlue}
        Exo & \checkmark & \checkmark & \textbf{73.4} & \textbf{82.0} & \textbf{76.2} & \textbf{65.3} & \textbf{74.2} \\
        \hline
    \end{tabular}}
    \label{tab:ablation_langdist_egotok}
\end{minipage}
\hfill
\begin{minipage}[t]{0.48\linewidth}
    \centering
    \setlength{\tabcolsep}{5pt}
    \caption{\textbf{Alternative \egotokensexpansion~Strategies.}}
    \vspace{-0.2cm}
    \resizebox{0.98\linewidth}{!}{
    \begin{tabular}{c|ccccc}
        \hline
        \multirow{2}{*}{\textbf{Token Strategy}} &
        \multirow{2}{*}{\shortstack{\textbf{Action}\\\textbf{Und.}}} &
        \multirow{2}{*}{\shortstack{\textbf{Task}\\\textbf{Regions}}} &
        \multirow{2}{*}{\textbf{HOI}} &
        \multirow{2}{*}{\shortstack{\textbf{Hand}\\\textbf{Ident.}}} &
        \multirow{2}{*}{\textbf{Avg.}} \\
        & & & & & \\
        \hline
        No tokens & 67.2 & 67.9 & 73.8 & 65.3 & 68.5 \\
        VE Self-attention & 68.2 & 69.2 & 74.0 & 68.3 & 69.9 \\
        Pre-trained CA & \textbf{74.3} & 79.9 & 74.1 & 65.1 & 73.4 \\
        \hline
        \rowcolor{LightBlue}
        \textbf{\modelname} & 73.4 & \textbf{82.0} & \textbf{76.2} & \textbf{65.3} & \textbf{74.2} \\
        \hline
    \end{tabular}}
    \label{tab:ablation_egotokens}
\end{minipage}
\vspace{-0.3cm}
\end{table*}

\subsection{Ablation and Analysis}
\label{sec:experiments_analysis}

\noindent\textbf{Alternate Viewpoint Transfer Strategies.}\quad
Table~\ref{tab:ablation_diststrat} compares our \langdistexpansion~against alternative viewpoint transfer strategies. In the \textit{Visual Feature Dist.} baseline, we train a VLM on egocentric videos and apply an MSE distillation loss between the pooled visual features of the frozen ego teacher and the exo student, analogous to the viewpoint transfer approach in Quattrocchi \textit{et al.}~\cite{quattrocchi2024_synch-all-you-need}.
In LLAVIDAL-style baseline, we train a VLM with ego-adaptive tokens on ego videos and directly pass these tokens as additional input to the language model when training the exo student, similar to LLAVIDAL\cite{reilly2025llavidal}. In EgoAda Tokens, we explore viewpoint transfer using the ego adaptive visual tokens themselves: we train an ego teacher VLM with additional tokens and perform feature-level distillation between the frozen ego tokens and exo tokens either after the visual encoder (Vis.), or after the language model (Lang.). We find that our proposed strategy surpasses all others, including the visual feature strategies used in prior non-VLM-based approaches (67.8\% vs 74.2\%), demonstrating the superior effectiveness of language-level knowledge transfer in VLMs. Notably, Exo2EgoVLM~\cite{zhang2025exo2egovlm}, designed for exo$\rightarrow$ego transfer, underperforms in the ego$\rightarrow$exo setting (70.3\% vs 74.2\%), indicating that methods effective for exo2ego do not directly translate to ego2exo transfer.

\noindent\textbf{Ablating Components.}\quad
Table~\ref{tab:ablation_langdist_egotok} ablates each component in \modelname.
Crucially, simply exposing the model to both ego and exo videos (Ego+Exo) yields negligible gains (+0.3\%). This demonstrates that viewpoint-conditioned semantics are not learned through shared training data alone and require explicit cross-view supervision.
In contrast, training on exo videos with sequence distillation alone improves average performance to 68.5\%. Ego-adaptive tokens alone yield stronger improvements at 72.2\%, particularly on the HOI task, indicating their effectiveness to enhance ego understanding. Combining both components achieves the best overall performance (74.2\% average), with consistent gains across all tasks, demonstrating the complementary nature of \langdistexpansion~and \egotokensexpansion. These results demonstrate that both modules are instrumental to enable ego-to-exo knowledge transfer.

\noindent\textbf{Alternative \egotokensexpansion~Strategies.}\quad
We explore alternative integration strategies for ego adaptive tokens in Table~\ref{tab:ablation_egotokens}. In VE Self-attention, the cross-attention matrices are removed and the tokens are directly appended to the visual features, thus they only interact with visual features through the self-attention of the vision encoder. In Pre-trained CA, we first train a VLM with ego-adaptive tokens on ego videos and initialize the student with the parameters of this model. VE Self-attention achieves modest gains compared to the VLM without tokens, but the lack of cross-attention limits the ability to learn robust token representations. While Pre-trained CA approaches \modelname's performance, we find it is less effective.

\begin{figure}[h]
    \centering

    \includegraphics[width=\linewidth]{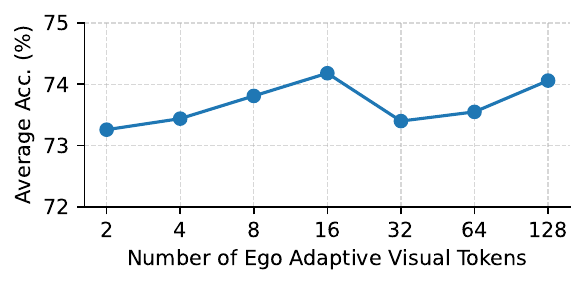}
    \vspace{-0.2cm}
    \caption{\textbf{Ablation on number of \egotokensexpansion.} Average accuracy on \benchname is reported.}
    \label{fig:analysis_numegotokens}
    \vspace{-0.2cm}
\end{figure}

\noindent\textbf{Analysis on number of Ego Adaptive Visual Tokens.}\quad
In Figure \ref{fig:analysis_numegotokens}, we present the accuracy of \modelname with varying numbers of \egotokensexpansion on \benchname. We find that our model's sensitivity to the number of tokens is low, and set the default to a value of $16$ tokens for all our experiments as a reasonable trade-off between efficiency and performance.

\begin{figure*}[t]
    \centering
    \begin{minipage}[t]{0.54\linewidth}
        \centering
        \includegraphics[width=\linewidth]{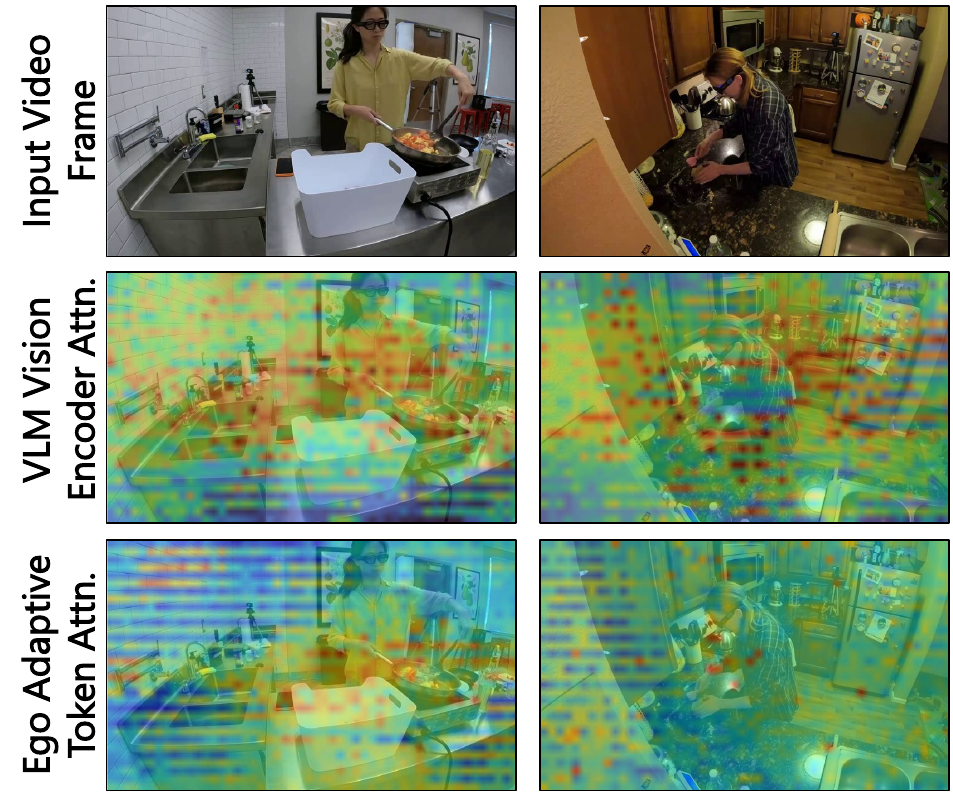}
        \caption{\textbf{Visualization of VLM Attention.} We compare the visual encoder's attention to attention of the ego-adaptive tokens.}
        \label{fig:analysis_attnvis}
    \end{minipage}
    \hfill
    \begin{minipage}[t]{0.44\linewidth}
        \centering
        \includegraphics[width=\linewidth]{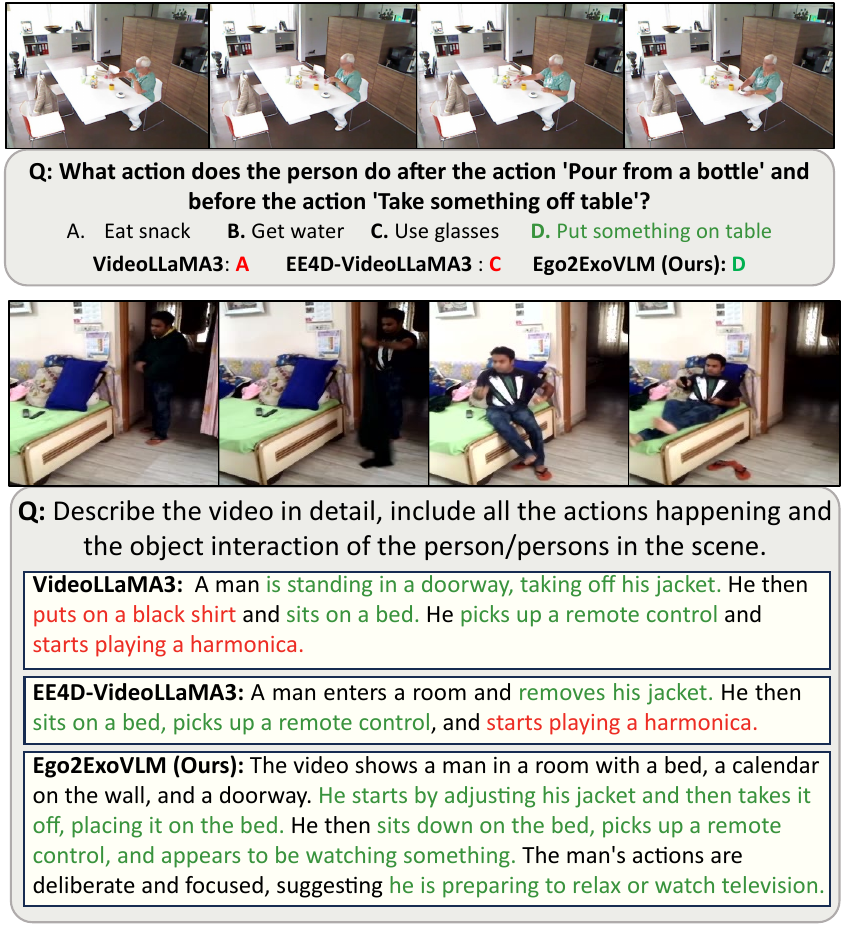}
        \caption{\textbf{Qualitative Results on the ADL-X dataset.} Additional qualitative results available in the Appendix.}
        \label{fig:qualitative_adlx}
    \end{minipage}
\end{figure*}

\noindent\textbf{Attention Visualization.}\quad
In Figure \ref{fig:analysis_attnvis}, we visualize the attentions of the visual encoder of an untrained student VLM, and of our \egotokensexpansion's to a video frames from two different videos. For the vision encoder, we visualize attention using attention rollout~\cite{abnar2020attentionrollout}, for our ego adaptive tokens, we visualize their attentions averaged across all tokens. We observe that the VLM vision encoder attends broadly to scene-level content, while our \egotokensexpansion~focus on localized regions where relevant interactions occur, such as human object interactions.

\noindent\textbf{Qualitative Results on ADL-X.}\quad Figure \ref{fig:qualitative_adlx} shows qualitative results of three VLMs: VideoLLaMA3, EE4D-VideoLLaMA3, and our proposed Ego2ExoVLM. We present example outputs of these VLMs on the Smarthome-AR, TSU-TC, and Charades Descriptions subsets of ADL-X. These results demonstrate that Ego2ExoVLM accurately identifies specific actions, object, and spatial details that are critical for ADL understanding.

\begin{table}[h]
    \renewcommand{\arraystretch}{1.2}
    \vspace{-0.2cm}
    \centering
    \caption{\textbf{Integration into different VLM backbone.}}
    \vspace{-0.2cm}
    \resizebox{0.99\linewidth}{!}{
    \begin{tabular}{ccccccc}
        \hline
        \multirow{2}{*}{\textbf{Model}} &
        \multirow{2}{*}{\textbf{Backbone}} &
        \multirow{2}{*}{\shortstack{\textbf{Action}\\\textbf{Und.}}} &
        \multirow{2}{*}{\shortstack{\textbf{Task}\\\textbf{Regions}}} &
        \multirow{2}{*}{\textbf{HOI}} &
        \multirow{2}{*}{\shortstack{\textbf{Hand}\\\textbf{Ident.}}} &
        \multirow{2}{*}{\textbf{Avg}} \\
        & & & & & & \\
        \hline
        Baseline & CLIP / Mistral & 30.1 & 32.6 & 30.6 & 54.6 & 37.0 \\
        Ego2ExoVLM & CLIP / Mistral & \textbf{41.2} & \textbf{43.8} & \textbf{40.9} & \textbf{56.8} & \textbf{45.7} \\
        \hline
    \end{tabular}}
    \vspace{-0.2cm}
    \label{tab:alternative_backbone}
\end{table}

\noindent\textbf{Robustness to Alternative VLM Backbones.}\quad
We investigate whether the improvements of \modelname generalize across different VLM backbones.
Table~\ref{tab:alternative_backbone} reports results using a CLIP-Large vision encoder and Mistral v0.2 language model, which represent a substantially weaker backbone compared to the SigLIP / Qwen2.5 backbone used in our main experiments. As expected, the absolute performance is lower due to the reduced model capacity and training data of the CLIP/Mistral backbone.
Nevertheless, \modelname consistently improves over the baseline across all tasks, demonstrating that the gains from ego-to-exo supervision are not specific to our backbone of choice.

\begin{table*}[t!]
\renewcommand{\arraystretch}{1.2}

\centering
\begin{minipage}[t]{0.42\linewidth}
    \centering
    \setlength{\tabcolsep}{5pt}
    \caption{\textbf{Perplexity and KL-Divergence with Teacher VLM.}}

    \resizebox{\linewidth}{!}{
    \begin{tabular}{c|cc}
        \hline
        \textbf{Method} & \textbf{Perplexity $\boldsymbol{(\downarrow)}$} & \textbf{KL-Divergence $\boldsymbol{(\downarrow)}$} \\
        \hline
        Untrained Student & 1.55 & 0.16 \\
        \hline
        \rowcolor{LightBlue}
        \textbf{\modelname} & \textbf{1.41} & \textbf{0.05} \\
        \hline
    \end{tabular}}
    \label{tab:analysis_perplexitykl}
\end{minipage}
\hfill
\begin{minipage}[t]{0.57\linewidth}
    \centering
    \setlength{\tabcolsep}{5pt}
    \caption{\textbf{\modelname's effectiveness on pseudo-paired Ego4D and HowTo100M.}}

    \resizebox{\linewidth}{!}{
    \begin{tabular}{ccccccc}
        \hline
        \textbf{Model} & \textbf{Training Data} & \textbf{Action Und.} & \textbf{Task Regions} & \textbf{HOI} & \textbf{Hand Ident.} & \textbf{Avg} \\
        \hline
        Baseline & Ego4D + HT100M & 65.3 & 66.8 & \textbf{75.1} & \textbf{65.3} & 68.1 \\
        \rowcolor{LightBlue}
        \textbf{\modelname} & Ego4D + HT100M & \textbf{69.5} & \textbf{77.6} & 71.7 & 62.8 & \textbf{70.4} \\
        \hline
    \end{tabular}}
    \label{tab:unpaired_data}
\end{minipage}

\end{table*}

\noindent\textbf{Perplexity and KL-Divergence with Teacher VLM.}\quad
In Table \ref{tab:analysis_perplexitykl}, we compute perplexity and KL-Divergence~\cite{kldivergence_information_theory} between the language outputs of the ego-observing teacher, an untrained student VLM, and \modelname on a subset of \benchname. Perplexity measures how closely a model's language output aligns with the teacher's, while KL-divergence measures the similarity between the logit-level outputs of each model. \modelname achieves lower scores on both metrics, indicating that the outputs of our \modelname are more closely aligned with those of the ego-observing teacher.

\section{Learning from Unpaired Ego-Exo Videos.}
\label{sec:unpaired_egoexo}

We next test whether exact temporal alignment is necessary. Instead of time-synced video pairs, we construct pseudo ego-exo pairs by retrieving exocentric clips from HowTo100M~\cite{miech2019howto100m} for each egocentric clip from Ego4D~\cite{ego4d} using the same text-based retrieval process as in \cite{xu2024retrievalaugmented}. These pairs originate from different videos and datasets, with no shared frame-level correspondence. Table~\ref{tab:unpaired_data} shows that training on these unpaired but semantically matched videos still improves over the baseline (Avg 70.4\% vs 68.1\%). This experiment demonstrates that \modelname’s language-mediated cross-view supervision does not require strictly time synchronized ego-exo video pairs.

\section{Conclusion}
\label{sec:conclusion}

We addressed the challenge of understanding egocentric properties from exocentric video observations, which remains difficult for existing VLMs despite their strong performance on general video understanding tasks. We presented \modelname, a VLM trained to infer egocentric properties from exocentric videos through time-synchronized cross-view knowledge transfer. To the best of our knowledge, \modelname is the first VLM to transfer knowledge from the egocentric to the exocentric viewpoint.
Our approach combines \langdistexpansion, which transfers egocentric reasoning through language-sequence supervision, with Ego-Adaptive Visual Tokens that enhance attention to ego-relevant spatial regions. To evaluate this capability, we introduced \benchname, a benchmark of 3.9K questions curated from the egocentric viewpoint while evaluated using exocentric videos. Through comprehensive evaluation, we demonstrate state-of-the-art performance on the ADL-X benchmark suite and strong improvements over baselines on \benchname, showing that egocentric supervision can effectively enable exocentric models to recover egocentric interaction cues. We further show that \modelname remains effective even when trained with non-time-synchronized ego-exo data, demonstrating that strict temporal alignment is not required for effective cross-view transfer.

\section{Acknowledgments}
This work was supported in part by the National Science Foundation (IIS-2245652) and the University of North Carolina at Charlotte. Computational resources were provided by the NSF National AI Research Resource Pilot (NAIRR-240338) and the NCShare initiative. We would like to thank Michael Ryoo and Razvan Bunescu for their valuable insights and discussions.
{
    \small
    \bibliographystyle{ieeenat_fullname}
    \bibliography{main}

@String(CVPR= {IEEE Conf. Comput. Vis. Pattern Recog.})

@String(ICCV= {Int. Conf. Comput. Vis.})

@String(ECCV= {Eur. Conf. Comput. Vis.})

@String(ICLR = {Int. Conf. Learn. Represent.})

@String(AAAI = {AAAI})

@misc{openai2025thinkwithimages,
  title={Thinking with Images},
  author={{OpenAI}},
  year={2025},
  month={April},
  url={https://openai.com/index/thinking-with-images/},
  note={Accessed: Jan 1 2026}
}

@article{Qwen2.5-VL,
  title={Qwen2.5-VL Technical Report},
  author={Bai, Shuai and Chen, Keqin and Liu, Xuejing and Wang, Jialin and Ge, Wenbin and Song, Sibo and Dang, Kai and Wang, Peng and Wang, Shijie and Tang, Jun and Zhong, Humen and Zhu, Yuanzhi and Yang, Mingkun and Li, Zhaohai and Wan, Jianqiang and Wang, Pengfei and Ding, Wei and Fu, Zheren and Xu, Yiheng and Ye, Jiabo and Zhang, Xi and Xie, Tianbao and Cheng, Zesen and Zhang, Hang and Yang, Zhibo and Xu, Haiyang and Lin, Junyang},
  journal={arXiv preprint arXiv:2502.13923},
  year={2025}
}

@article{damonlpsg2025videollama3,
  title={VideoLLaMA 3: Frontier Multimodal Foundation Models for Image and Video Understanding},
  author={Zhang, Boqiang and Li, Kehan and Cheng, Zesen and Hu, Zhiqiang and Yuan, Yuqian and Chen, Guanzheng and Leng, Sicong and Jiang, Yuming and Zhang, Hang and Li, Xin and Jin, Peng and Zhang, Wenqi and Wang, Fan and Bing, Lidong and Zhao, Deli},
  journal={arXiv preprint arXiv:2501.13106},
  year={2025},
  url = {https://arxiv.org/abs/2501.13106}
}

@misc{xue2025blip3,
      title={xGen-MM (BLIP-3): A Family of Open Large Multimodal Models},
      author={Xue, Le and Shu, Manli and Awadalla, Anas and Wang, Jun and Yan, An and Purushwalkam, Senthil and Zhou, Honglu and Prabhu, Viraj and Dai, Yutong and Ryoo, Michael S and Kendre, Shrikant and Zhang, Jieyu and Tseng, Shaoyen and Lujan-Moreno, Gustavo A and Olson, Matthew L and Hinck, Musashi and Cobbley, David and Lal, Vasudev and Qin, Can and Zhang, Shu and Chen, Chia-Chih and Yu, Ning and Tan, Juntao and Awalgaonkar, Tulika Manoj and Heinecke, Shelby and Wang, Huan and Choi, Yejin and Schmidt, Ludwig and Chen, Zeyuan and Savarese, Silvio and Niebles, Juan Carlos and Xiong, Caiming and Xu, Ran},
      year={2025},
      eprint={2408.08872},
      archivePrefix={arXiv}
}

@inproceedings{
socratic,
title={Socratic Models: Composing Zero-Shot Multimodal Reasoning with Language},
author={Andy Zeng and Maria Attarian and brian ichter and Krzysztof Marcin Choromanski and Adrian Wong and Stefan Welker and Federico Tombari and Aveek Purohit and Michael S Ryoo and Vikas Sindhwani and Johnny Lee and Vincent Vanhoucke and Pete Florence},
booktitle={The Eleventh International Conference on Learning Representations },
year={2023},
url={https://openreview.net/forum?id=G2Q2Mh3avow}
}

@inproceedings{mmbench,
      title={MMBench: Is Your Multi-modal Model an All-around Player?},
      author={Liu, Yuanzhan and Duan, Haodong and Zhang, Yuanhan and Li, Bo and Zhang, Songyang and Zhao, Wangbo and Yuan, Yike and Wang, Jiaqi and He, Conghui and Liu, Ziwei and Chen, Kai and Lin, Dahua},
      booktitle={European Conference on Computer Vision},
      year={2024}
}

@article{lai2023lisa,
  title={LISA: Reasoning Segmentation via Large Language Model},
  author={Lai, Xin and Tian, Zhuotao and Chen, Yukang and Li, Yanwei and Yuan, Yuhui and Liu, Shu and Jia, Jiaya},
  journal={arXiv preprint arXiv:2308.00692},
  year={2023}
}

@article{Ranasinghe2024LearningTL,
  title={Learning to Localize Objects Improves Spatial Reasoning in Visual-LLMs},
  author={Kanchana Ranasinghe and Satya Narayan Shukla and Omid Poursaeed and Michael S. Ryoo and Tsung-Yu Lin},
  journal={2024 IEEE/CVF Conference on Computer Vision and Pattern Recognition (CVPR)},
  year={2024},
  pages={12977-12987},
  url={https://api.semanticscholar.org/CorpusID:269043025}
}

@misc{qwen2025qwen25technicalreport,
      title={Qwen2.5 Technical Report}, 
      author={{Qwen Team} and Yang, An and Yang, Baosong and Zhang, Beichen and Hui, Binyuan and Zheng, Bo and Yu, Bowen and Li, Chengyuan and Liu, Dayiheng and Huang, Fei and Wei, Haoran and Lin, Huan and Yang, Jian and Tu, Jianhong and Zhang, Jianwei and Yang, Jianxin and Yang, Jiaxi and Zhou, Jingren and Lin, Junyang and Dang, Kai and Lu, Keming and Bao, Keqin and Yang, Kexin and Yu, Le and Li, Mei and Xue, Mingfeng and Zhang, Pei and Zhu, Qin and Men, Rui and Lin, Runji and Li, Tianhao and Tang, Tianyi and Xia, Tingyu and Ren, Xingzhang and Ren, Xuancheng and Fan, Yang and Su, Yang and Zhang, Yichang and Wan, Yu and Liu, Yuqiong and Cui, Zeyu and Zhang, Zhenru and Qiu, Zihan},
      year={2025},
      eprint={2412.15115},
      archivePrefix={arXiv},
      primaryClass={cs.CL},
      url={https://arxiv.org/abs/2412.15115}, 
}

@misc{meta2024llama3herdmodels,
      title={The Llama 3 Herd of Models},
      author={Meta},
      year={2024},
      eprint={2407.21783},
      archivePrefix={arXiv}
}

@misc{radford2021openaiclip,
      title={Learning Transferable Visual Models From Natural Language Supervision},
      author={Radford, Alec and Kim, Jong Wook and Hallacy, Chris and Ramesh, Aditya and Goh, Gabriel and Agarwal, Sandhini and Sastry, Girish and Askell, Amanda and Mishkin, Pamela and Clark, Jack and Krueger, Gretchen and Sutskever, Ilya},
      year={2021},
      eprint={2103.00020},
      archivePrefix={arXiv}
}

@inproceedings{zhai2023siglipv1,
      title={Sigmoid Loss for Language Image Pre-Training}, 
      author={Zhai, Xiaohua and Mustafa, Basil and Kolesnikov, Alexander and Beyer, Lucas},
      booktitle={Proceedings of the IEEE/CVF International Conference on Computer Vision},
      year={2023},
      pages={}, 
      organization={IEEE}
}

@misc{zhang2024llavavideo,
    title     = {Video Instruction Tuning With Synthetic Data}, 
    author    = {Zhang, Yuanhan and Wu, Jinming and Li, Wei and Li, Bo and Ma, Zejun and Liu, Ziwei and Li, Chunyuan},
    year      = {2024},
    eprint    = {2410.02713},
    archivePrefix = {arXiv},
    primaryClass  = {cs.CV},
    url       = {https://arxiv.org/abs/2410.02713}
}

@inproceedings{chen2024sharegpt4video,
      title={ShareGPT4Video: Improving Video Understanding and Generation with Better Captions},
      author={Chen, Lin and Wei, Xilin and Li, Jinsong and Dong, Xiaoyi and Zhang, Pan and Zang, Yuhang and Chen, Zehui and Duan, Haodong and Lin, Bin and Tang, Zhenyu and Yuan, Li and Qiao, Yu and Lin, Dahua and Zhao, Feng and Wang, Jiaqi},
      booktitle={Advances in Neural Information Processing Systems},
      year={2024}
}

@misc{maaz2024videogptplus,
      title={VideoGPT+: Integrating Image and Video Encoders for Enhanced Video Understanding},
      author={Maaz, Muhammad and Rasheed, Hanoona and Khan, Salman and Khan, Fahad},
      year={2024},
      eprint={2406.09418},
      archivePrefix={arXiv}
}

@article{he2026bridgingperspectives,
  title   = {Bridging Perspectives: A Survey on Cross-view Collaborative Intelligence with Egocentric-Exocentric Vision},
  author  = {Yuping He and Yifei Huang and Guo Chen and Lidong Lu and Baoqi Pei and Jilan Xu and Tong Lu and Yoichi Sato},
  journal = {International Journal of Computer Vision},
  volume  = {134},
  pages   = {62},
  year    = {2026}
}

@inproceedings{cvpr2025egoexo4d,
      title={Ego-Exo4D: Understanding Skilled Human Activity from First- and Third-Person Perspectives}, 
      author={Grauman, Kristen and Westbury, Andrew and Torresani, Lorenzo and Kitani, Kris and Malik, Jitendra and Afouras, Triantafyllos and Ashutosh, Kumar and Baiyya, Vijay and Bansal, Siddhant and Boote, Bikram and Byrne, Eugene and Chavis, Zach and Chen, Joya and Cheng, Feng and Chu, Fu-Jen and Crane, Sean and Dasgupta, Avijit and Dong, Jing and Escobar, Maria and Forigua, Cristhian and Gebreselasie, Abrham and Haresh, Sanjay and Huang, Jing and Islam, Md Mohaiminul and Jain, Suyog and Khirodkar, Rawal and Kukreja, Devansh and Liang, Kevin J and Liu, Jia-Wei and Majumder, Sagnik and Mao, Yongsen and Martin, Miguel and Mavroudi, Effrosyni and Nagarajan, Tushar and Ragusa, Francesco and Ramakrishnan, Santhosh Kumar and Seminara, Luigi and Somayazulu, Arjun and Song, Yale and Su, Shan and Xue, Zihui and Zhang, Edward and Zhang, Jinxu and Castillo, Angela and Chen, Changan and Fu, Xinzhu and Furuta, Ryosuke and Gonzalez, Cristina and Gupta, Prince and Hu, Jiabo and Huang, Yifei and Huang, Yiming and Khoo, Weslie and Kumar, Anush and Kuo, Robert and Lakhavani, Sach and Liu, Miao and Luo, Mi and Luo, Zhengyi and Meredith, Brighid and Miller, Austin and Oguntola, Oluwatumininu and Pan, Xiaqing and Peng, Penny and Pramanick, Shraman and Ramazanova, Merey and Ryan, Fiona and Shan, Wei and Somasundaram, Kiran and Song, Chenan and Southerland, Audrey and Tateno, Masatoshi and Wang, Huiyu and Wang, Yuchen and Yagi, Takuma and Yan, Mingfei and Yang, Xitong and Yu, Zecheng and Zha, Shengxin Cindy and Zhao, Chen and Zhao, Ziwei and Zhu, Zhifan and Zhuo, Jeff and Arbelaez, Pablo and Bertasius, Gedas and Crandall, David and Damen, Dima and Engel, Jakob and Farinella, Giovanni Maria and Furnari, Antonino and Ghanem, Bernard and Hoffman, Judy and Jawahar, C. V. and Newcombe, Richard and Park, Hyun Soo and Rehg, James M. and Sato, Yoichi and Savva, Manolis and Shi, Jianbo and Shou, Mike Zheng and Wray, Michael},
      booktitle={Proceedings of the IEEE/CVF Conference on Computer Vision and Pattern Recognition},
      year={2025}
}

@article{sigurdsson2018charades-ego,
  title   = {Charades-ego: A large-scale dataset of paired third and first person videos},
  author  = {Sigurdsson, Gunnar A and Gupta, Abhinav and Schmid, Cordelia and Farhadi, Ali and Alahari, Karteek},
  journal = {arXiv preprint arXiv:1804.09626},
  year    = {2018}
}

@inproceedings{reilly2025llavidal,
  title        = {LLAVIDAL: A Large Language-Vision Model for Daily Activities of Living},
  author       = {Reilly, Dominick and Chakraborty, Rajatsubhra and Sinha, Arkaprava and Govind, Manish Kumar and Wang, Pu and Bremond, Francois and Xue, Le and Das, Srijan},
  booktitle    = {Proceedings of the IEEE/CVF Conference on Computer Vision and Pattern Recognition},
  year         = {2025}
}

@article{liu2019ntu,
title = {{NTU RGB+D 120: A Large-Scale Benchmark for 3D Human Activity Understanding}},
author = {Liu, Jun and Shahroudy, Amir and Perez, Mauricio and Wang, Gang and Duan, Ling-Yu and Kot, Alex C.},
journal = {IEEE Transactions on Pattern Analysis and Machine Intelligence},
year = {2019}
}

@inproceedings{jang2020etri,
title = {{ETRI-Activity3D: A Large-Scale RGB-D Dataset for Robots to Recognize Daily Activities of the Elderly}},
author = {Jang, Jinhyeok and Kim, Dohyung and Park, Cheonshu and Jang, Minsu and Lee, Jaeyeon and Kim, Jaehong},
booktitle = {IEEE/RSJ International Conference on Intelligent Robots and Systems (IROS)},
year = {2020}
}

@inproceedings{das2019toyotasmarthome,
  title        = {Toyota Smarthome: Real-World Activities of Daily Living},
  author       = {Das, Srijan and Dai, Rui and Koperski, Michal and Minciullo, Luca and Garattoni, Lorenzo and Bremond, Francois and Francesca, Gianpiero},
  booktitle    = {Proceedings of the IEEE/CVF International Conference on Computer Vision},
  year         = {2019},
  pages        = {833--842}
}

@inproceedings{liu2023_llava,
      title={Visual Instruction Tuning}, 
      author={Liu, Haotian and Li, Chunyuan and Wu, Qingyang and Lee, Yong Jae},
      booktitle={Advances in Neural Information Processing Systems (NeurIPS)},
      year={2023},
}

@InProceedings{quattrocchi2024_synch-all-you-need,
 title={Synchronization is All You Need: Exocentric-to-Egocentric Transfer for Temporal Action Segmentation with Unlabeled Synchronized Video Pairs},
 author={Quattrocchi, Camillo and Furnari, Antonino and Di Mauro, Daniele and Giuffrida, Mario Valerio and Farinella, Giovanni Maria},
 booktitle={European Conference on Computer Vision (ECCV)},
 year={2024}
}

@inproceedings{reilly2026viscop,
  title     = {VisCoP: Visual Probing for Video Domain Adaptation of Vision Language Models},
  author    = {Dominick Reilly and Manish Kumar Govind and Le Xue and Srijan Das},
  booktitle = {Proceedings of the European Conference on Computer Vision (ECCV)},
  year      = {2026}
}

@inproceedings{luo2025viewpointrosetta,
      title={Viewpoint Rosetta Stone: Unlocking Unpaired Ego-Exo Videos for View-invariant Representation Learning},
      author={Luo, Mi and Xue, Zihui and Dimakis, Alex and Grauman, Kristen},
      booktitle={Proceedings of the IEEE/CVF Conference on Computer Vision and Pattern Recognition},
      year={2025}
    }

@inproceedings{ohkawa2023_exo2egodvc,
      title={Exo2EgoDVC: Dense Video Captioning of Egocentric Procedural Activities Using Web Instructional Videos}, 
      author={Takehiko Ohkawa and Takuma Yagi and Taichi Nishimura and Ryosuke Furuta and Atsushi Hashimoto and Yoshitaka Ushiku and Yoichi Sato},
      booktitle={Proceedings of the IEEE/CVF Winter Conference on Applications of Computer Vision (WACV)},
      year={2025}
}

@inproceedings{xu2024retrievalaugmented,
  title={Retrieval-augmented egocentric video captioning},
  author={Xu, Jilan and Huang, Yifei and Hou, Junlin and Chen, Guo and Zhang, Yuejie and Feng, Rui and Xie, Weidi},
  booktitle={Proceedings of the IEEE/CVF Conference on Computer Vision and Pattern Recognition},
  year={2024}
}

@misc{EMBED,
      title={Unlocking Exocentric Video-Language Data for Egocentric Video Representation Learning}, 
      author={Zi-Yi Dou and Xitong Yang and Tushar Nagarajan and Huiyu Wang and Jing Huang and Nanyun Peng and Kris Kitani and Fu-Jen Chu},
      year={2024},
      eprint={2408.03567},
      archivePrefix={arXiv},
      primaryClass={cs.CV},
      url={https://arxiv.org/abs/2408.03567}, 
}

@article{llama,
  title={LLaMA: Open and Efficient Foundation Language Models},
  author={Hugo Touvron and Thibaut Lavril and Gautier Izacard and Xavier Martinet and Marie-Anne Lachaux and Timoth{\'e}e Lacroix and Baptiste Rozi{\`e}re and Naman Goyal and Eric Hambro and Faisal Azhar and Aurelien Rodriguez and Armand Joulin and Edouard Grave and Guillaume Lample},
  journal={ArXiv},
  year={2023},
  volume={abs/2302.13971},
  url={https://api.semanticscholar.org/CorpusID:257219404}
}

@inproceedings{gpt,
  title={Language Models are Few-Shot Learners},
  author={Tom B. Brown and Benjamin Mann and Nick Ryder and Melanie Subbiah and Jared Kaplan and Prafulla Dhariwal and Arvind Neelakantan and Pranav Shyam and Girish Sastry and Amanda Askell and Sandhini Agarwal and Ariel Herbert-Voss and Gretchen Krueger and Tom Henighan and Rewon Child and Aditya Ramesh and Daniel M. Ziegler and Jeff Wu and Clemens Winter and Christopher Hesse and Mark Chen and Eric Sigler and Mateusz Litwin and Scott Gray and Benjamin Chess and Jack Clark and Christopher Berner and Sam McCandlish and Alec Radford and Ilya Sutskever and Dario Amodei},
  booktitle={Advances in Neural 
Information Processing Systems (NeurIPS)},
  year={2020}
}

@misc{vicuna2023,
    title = {Vicuna: An Open-Source Chatbot Impressing GPT-4 with 90\%* ChatGPT Quality},
    url = {https://lmsys.org/blog/2023-03-30-vicuna/},
    author = {Chiang, Wei-Lin and Li, Zhuohan and Lin, Zi and Sheng, Ying and Wu, Zhanghao and Zhang, Hao and Zheng, Lianmin and Zhuang, Siyuan and Zhuang, Yonghao and Gonzalez, Joseph E. and Stoica, Ion and Xing, Eric P.},
    month = {March},
    year = {2023}
}

@misc{li2024_llava-next-interleave,
      title={LLaVA-NeXT-Interleave: Tackling Multi-image, Video, and 3D in Large Multimodal Models}, 
      author={Feng Li and Renrui Zhang and Hao Zhang and Yuanhan Zhang and Bo Li and Wei Li and Zejun Ma and Chunyuan Li},
      year={2024},
      eprint={2407.07895},
      archivePrefix={arXiv},
      primaryClass={cs.CV},
      url={https://arxiv.org/abs/2407.07895}, 
}

@article{zhang2024_llava-hound-dpo,
      title={Direct Preference Optimization of Video Large Multimodal Models from Language Model Reward}, 
      author={Ruohong Zhang and Liangke Gui and Zhiqing Sun and Yihao Feng and Keyang Xu and Yuanhan Zhang and Di Fu and Chunyuan Li and Alexander Hauptmann and Yonatan Bisk and Yiming Yang},
      year={2024},
      journal={ArXiv arXiv:2404.01258}
}

@article{xu2024_slowfast-llava,
      title={SlowFast-LLaVA: A Strong Training-Free Baseline for Video Large Language Models}, 
      author={Mingze Xu and Mingfei Gao and Zhe Gan and Hong-You Chen and Zhengfeng Lai and Haiming Gang and Kai Kang and Afshin Dehghan},
      year={2024},
      journal={ArXiv preprint arXiv:2407.15841}
}

@inproceedings{videollava,
  title={Video-LLaVA: Learning United Visual Representation by Alignment Before Projection},
  author={Lin, Bin and Zhu, Bin and Ye, Yang and Ning, Munan and Jin, Peng and Yuan, Li},
  booktitle={Conference on Empirical Methods in Natural Language Processing (EMNLP)},
  year={2024}
}

@inproceedings{videollama,
  title={Video-LLaMA: An Instruction-tuned Audio-Visual Language Model for Video Understanding},
  author={Hang Zhang and Xin Li and Lidong Bing},
  journal={ArXiv},
  booktitle={Conference on Empirical Methods in Natural Language Processing (EMNLP)},
  year={2023}
}

@inproceedings{videochatgpt,
  title={Video-ChatGPT: Towards Detailed Video Understanding via Large Vision and Language Models},
  author={Muhammad Maaz and Hanoona Abdul Rasheed and Salman H. Khan and Fahad Shahbaz Khan},
  booktitle={Annual Meeting of the Association for Computational Linguistics (ACL 2024)},
  year={2024}
}

@inproceedings{jin2023chatunivi,
  title={Chat-UniVi: Unified Visual Representation Empowers Large Language Models with Image and Video Understanding}, 
  author={Peng Jin and Ryuichi Takanobu and Caiwan Zhang and Xiaochun Cao and Li Yuan},
  booktitle={Proceedings of the IEEE/CVF Conference on Computer Vision and Pattern Recognition (CVPR)},
  year={2024}
}

@article{li2024_llava-onevision,
      title={LLaVA-OneVision: Easy Visual Task Transfer}, 
      author={Bo Li and Yuanhan Zhang and Dong Guo and Renrui Zhang and Feng Li and Hao Zhang and Kaichen Zhang and Peiyuan Zhang and Yanwei Li and Ziwei Liu and Chunyuan Li},
      year={2024},
      journal={ArXiv preprint arXiv:2408.03326}
}

@InProceedings{ego4d,
    author    = {Grauman, Kristen and Westbury, Andrew and Byrne, Eugene and Chavis, Zachary and Furnari, Antonino and Girdhar, Rohit and Hamburger, Jackson and Jiang, Hao and Liu, Miao and Liu, Xingyu and Martin, Miguel and Nagarajan, Tushar and Radosavovic, Ilija and Ramakrishnan, Santhosh Kumar and Ryan, Fiona and Sharma, Jayant and Wray, Michael and Xu, Mengmeng and Xu, Eric Zhongcong and Zhao, Chen and Bansal, Siddhant and Batra, Dhruv and Cartillier, Vincent and Crane, Sean and Do, Tien and Doulaty, Morrie and Erapalli, Akshay and Feichtenhofer, Christoph and Fragomeni, Adriano and Fu, Qichen and Gebreselasie, Abrham and Gonz\'alez, Cristina and Hillis, James and Huang, Xuhua and Huang, Yifei and Jia, Wenqi and Khoo, Weslie and Kol\'a\v{r}, J\'achym and Kottur, Satwik and Kumar, Anurag and Landini, Federico and Li, Chao and Li, Yanghao and Li, Zhenqiang and Mangalam, Karttikeya and Modhugu, Raghava and Munro, Jonathan and Murrell, Tullie and Nishiyasu, Takumi and Price, Will and Ruiz, Paola and Ramazanova, Merey and Sari, Leda and Somasundaram, Kiran and Southerland, Audrey and Sugano, Yusuke and Tao, Ruijie and Vo, Minh and Wang, Yuchen and Wu, Xindi and Yagi, Takuma and Zhao, Ziwei and Zhu, Yunyi and Arbel\'aez, Pablo and Crandall, David and Damen, Dima and Farinella, Giovanni Maria and Fuegen, Christian and Ghanem, Bernard and Ithapu, Vamsi Krishna and Jawahar, C. V. and Joo, Hanbyul and Kitani, Kris and Li, Haizhou and Newcombe, Richard and Oliva, Aude and Park, Hyun Soo and Rehg, James M. and Sato, Yoichi and Shi, Jianbo and Shou, Mike Zheng and Torralba, Antonio and Torresani, Lorenzo and Yan, Mingfei and Malik, Jitendra},
    title     = {Ego4D: Around the World in 3,000 Hours of Egocentric Video},
    booktitle = {Proceedings of the IEEE/CVF Conference on Computer Vision and Pattern Recognition (CVPR)},
    month     = {June},
    year      = {2022},
    pages     = {18995-19012}
}

@article{thatipelli2024_exo2ego-survey,
      title={Exocentric To Egocentric Transfer For Action Recognition: A Short Survey}, 
      author={Anirudh Thatipelli and Shao-Yuan Lo and Amit K. Roy-Chowdhury},
      year={2024},
      journal={ArXiv preprint arXiv:2410.20621}
}

@InProceedings{xu2018_egoexo-reid,
  title={Joint Person Segmentation and Identification in Synchronized First- and Third-person Videos},
  author={Xu, Mingze and Fan, Chenyou and Wang, Yuchen and Ryoo, Michael S. and Crandall, David J.},
  booktitle={European Conference on Computer Vision (ECCV)},
  pages={674--693},
  year={2018}
}

@InProceedings{sigurdsson2018_actor-observer,
  title={Actor and Observer: Joint Modeling of First and Third-Person Videos},
  author={Sigurdsson, Gunnar A. and Gupta, Abhinav and Schmid, Cordelia and Farhadi, Ali and Alahari, Karteek},
  booktitle={Proceedings of the IEEE Conference on Computer Vision and Pattern Recognition (CVPR)},
  pages={7396--7404},
  year={2018}
}

@InProceedings{yu2019_egoexo-joint-attention,
 title={What I See Is What You See: Joint Attention Learning for First and Third Person Video Co-analysis},
 author={Yu, Huangyue and Cai, Minjie and Liu, Yunfei and Lu, Feng},
 booktitle={Proceedings of the 27th ACM International Conference on Multimedia (MM)},
 pages={1926--1934},
 year={2019}
}

@InProceedings{xue2023_egoexo-ae2,
      title={Learning Fine-grained View-Invariant Representations from Unpaired Ego-Exo Videos via Temporal Alignment},
      author={Xue, Zihui and Grauman, Kristen},
      booktitle={Advances in Neural Information 
Processing Systems (NeurIPS)},
      year={2023}
}

@InProceedings{ardeshir2018_egoexo,
title = {An exocentric look at egocentric actions and vice versa},
booktitle = {Computer Vision and Image Understanding},
volume = {171},
pages = {61-68},
year = {2018},
author = {Shervin Ardeshir and Ali Borji}
}

@InProceedings{li2021_egoexo-transfer,
 title={Ego-Exo: Transferring Visual Representations from Third-person to First-person Videos},
 author={Li, Yanghao and Nagarajan, Tushar and Xiong, Bo and Grauman, Kristen},
 booktitle={Proceedings of the IEEE/CVF Conference on Computer Vision and Pattern Recognition (CVPR)},
 pages={10995--11005},
 year={2021}
}

@InProceedings{xu2023_egoexo-POV,
 title={POV: Prompt-Oriented View-Agnostic Learning for Egocentric Hand-Object Interaction in the Multi-view World},
 author={Xu, Boshen and Zheng, Sipeng and Jin, Qin},
 booktitle={Proceedings of the 31st ACM International Conference on Multimedia (MM)},
 pages={2807--2816},
 year={2023}
}

@InProceedings{wang2023_egoexo-SUML,
 title={Learning from Semantic Alignment between Unpaired Multiviews for Egocentric Video Recognition},
 author={Wang, Qitong and Zhao, Long and Yuan, Liangzhe and Liu, Ting and Peng, Xi},
 booktitle={Proceedings of the IEEE/CVF International Conference on Computer Vision (ICCV)},
 pages={20453--20463},
 year={2023}
}

@InProceedings{rai2024_egoexo-objectaffordance,
    author    = {Rai, Arushi and Buettner, Kyle and Kovashka, Adriana},
    title     = {Strategies to Leverage Foundational Model Knowledge in Object Affordance Grounding},
    booktitle = {Proceedings of the IEEE/CVF Conference on Computer Vision and Pattern Recognition (CVPR) Workshops},
    month     = {June},
    year      = {2024},
    pages     = {1714-1723}
}

@inproceedings{park2026egoworld,
  title     = {EgoWorld: Translating Exocentric View to Egocentric View using Rich Exocentric Observations},
  author    = {Junho Park and Andrew Sangwoo Ye and Taein Kwon},
  booktitle = {International Conference on Learning Representations (ICLR)},
  year      = {2026}
}

@inproceedings{luo2024exo2ego,
  title     = {Put Myself in Your Shoes: Lifting the Egocentric Perspective from Exocentric Videos},
  author    = {Mi Luo and Zihui Xue and Alex Dimakis and Kristen Grauman},
  booktitle = {Proceedings of the European Conference on Computer Vision (ECCV)},
  year      = {2024}
}

@inproceedings{mahdi2026syn2seq,
  title     = {From Synchrony to Sequence: Exo-to-Ego Generation via Interpolation},
  author    = {Mohammad Mahdi and Nedko Savov and Danda Pani Paudel and Luc Van Gool},
  booktitle = {Proceedings of the European Conference on Computer Vision (ECCV)},
  year      = {2026}
}

@INPROCEEDINGS{Damen2018EPICKITCHENS,
   title={Scaling Egocentric Vision: The EPIC-KITCHENS Dataset},
   author={Damen, Dima and Doughty, Hazel and Farinella, Giovanni Maria  and Fidler, Sanja and 
           Furnari, Antonino and Kazakos, Evangelos and Moltisanti, Davide and Munro, Jonathan 
           and Perrett, Toby and Price, Will and Wray, Michael},
   booktitle={European Conference on Computer Vision (ECCV)},
   year={2018}
}

@article{kinetics600,
  author     = {Jo{\~{a}}o Carreira and
                Eric Noland and
                Andras Banki{-}Horvath and
                Chloe Hillier and
                Andrew Zisserman},
  title      = {A Short Note about Kinetics-600},
  journal    = {CoRR},
  volume     = {abs/1808.01340},
  year       = {2018},
  url        = {http://arxiv.org/abs/1808.01340},
  eprinttype = {arXiv},
  eprint     = {1808.01340},
  bibsource  = {dblp computer science bibliography, https://dblp.org}
}

@inproceedings{fu2025videomme,
  title        = {Video-MME: The First-Ever Comprehensive Evaluation Benchmark of Multi-modal Large Language Models in Video Analysis},
  author       = {Fu, Chaoyou and Dai, Yuhan and Luo, Yongdong and Li, Lei and Ren, Shuhuai and Zhang, Renrui and Wang, Zihan and Zhou, Chenyu and Shen, Yunhang and Zhang, Mengdan and Chen, Peixian and Li, Yanwei and Lin, Shaohui and Zhao, Sirui and Li, Ke and Xu, Tong and Zheng, Xiawu and Chen, Enhong and Shan, Caifeng and He, Ran and Sun, Xing},
  booktitle    = {Proceedings of the IEEE/CVF Conference on Computer Vision and Pattern Recognition},
  year         = {2025}
}

@article{kevin2022egovlp,
  title={Egocentric Video-Language Pretraining},
  author={Lin, Kevin Qinghong and Wang, Alex Jinpeng and Soldan, Mattia and Wray, Michael and Yan, Rui and Xu, Eric Zhongcong and Gao, Difei and Tu, Rongcheng and Zhao, Wenzhe and Kong, Weijie and others},
  journal={arXiv preprint arXiv:2206.01670},
  year={2022}
}

@inproceedings{egoschema,
title={EgoSchema: A Diagnostic Benchmark for Very Long-form Video Language Understanding},
author={Karttikeya Mangalam and Raiymbek Akshulakov and Jitendra Malik},
booktitle={Thirty-seventh Conference on Neural Information Processing Systems Datasets and Benchmarks Track},
year={2023},
url={https://openreview.net/forum?id=JVlWseddak}
}

@inproceedings{ye2025mmego,
  title        = {MM-Ego: Towards Building Egocentric Multimodal LLMs for Video QA},
  author       = {Hanrong Ye and Haotian Zhang and Erik Daxberger and Lin Chen and Zongyu Lin and Yanghao Li and Bowen Zhang and Haoxuan You and Dan Xu and Zhe Gan and Jiasen Lu and Yinfei Yang},
  booktitle    = {Proceedings of the International Conference on Learning Representations (ICLR)},
  year         = {2025}
}

@inproceedings{lemma,
author={Jia, Baoxiong and Chen, Yixin and Huang, Siyuan and Zhu, Yixin and Zhu, Song-Chun}, 
title={LEMMA: A Multiview Dataset for Learning Multi-agent Multi-view Activities}, 
booktitle={Proceedings of the European Conference on Computer Vision (ECCV)}, 
year={2020}
}

@inproceedings{huang2024egoexolearn,
  title        = {EgoExoLearn: A Dataset for Bridging Asynchronous Ego- and Exo-centric View of Procedural Activities in Real World},
  author       = {Yifei Huang and Guo Chen and Jilan Xu and Mingfang Zhang and Lijin Yang and Baoqi Pei and Hongjie Zhang and Lu Dong and Yali Wang and Limin Wang and Yu Qiao},
  booktitle    = {Proceedings of the IEEE/CVF Conference on Computer Vision and Pattern Recognition (CVPR)},
  year         = {2024}
}

@article{he2025egoexobench,
  title        = {EgoExoBench: A Benchmark for First- and Third-person View Video Understanding in MLLMs},
  author       = {Yuping He and Yifei Huang and Guo Chen and Baoqi Pei and Jilan Xu and Tong Lu and Jiangmiao Pang},
  journal      = {arXiv preprint arXiv:2507.18342},
  year         = {2025}
}

@article{damonlpsg2024videollama2,
  title={VideoLLaMA 2: Advancing Spatial-Temporal Modeling and Audio Understanding in Video-LLMs},
  author={Cheng, Zesen and Leng, Sicong and Zhang, Hang and Xin, Yifei and Li, Xin and Chen, Guanzheng and Zhu, Yongxin and Zhang, Wenqi and Luo, Ziyang and Zhao, Deli and Bing, Lidong},
  journal={arXiv preprint arXiv:2406.07476},
  year={2024},
  url = {https://arxiv.org/abs/2406.07476}
}

@article{wang2024qwen2vl,
  title   = {Qwen2-VL: Enhancing Vision-Language Model's Perception of the World at Any Resolution},
  author  = {Peng Wang and Shuai Bai and Sinan Tan and Shijie Wang and Zhihao Fan and Jinze Bai and Keqin Chen and Xuejing Liu and Jialin Wang and Wenbin Ge and Yang Fan and Kai Dang and Mengfei Du and Xuancheng Ren and Rui Men and Dayiheng Liu and Chang Zhou and Jingren Zhou and Junyang Lin},
  journal = {arXiv preprint arXiv:2409.12191},
  year    = {2024}
}

@InProceedings{hong2023cogagent,
      title={CogAgent: A Visual Language Model for GUI Agents}, 
      author={Wenyi Hong and Weihan Wang and Qingsong Lv and Jiazheng Xu and Wenmeng Yu and Junhui Ji and Yan Wang and Zihan Wang and Yuxiao Dong and Ming Ding and Jie Tang},
      booktitle={Proceedings of the IEEE/CVF Conference on Computer Vision and Pattern Recognition (CVPR)},
      year={2024}
}

@misc{openai2024gpt4ocard,
      title={GPT-4o System Card}, 
      author={OpenAI},
      year={2024},
      eprint={2410.21276},
      archivePrefix={arXiv},
      primaryClass={cs.CL},
      url={https://arxiv.org/abs/2410.21276}, 
}

@inproceedings{cogvlm,
  title={CogVLM: Visual Expert for Pretrained Language Models},
  author={Weihan Wang and Qingsong Lv and Wenmeng Yu and Wenyi Hong and Ji Qi and Yan Wang and Junhui Ji and Zhuoyi Yang and Lei Zhao and Xixuan Song and Jiazheng Xu and Bin Xu and Juanzi Li and Yuxiao Dong and Ming Ding and Jie Tang},
  booktitle={Advances in Neural Information 
Processing Systems (NeurIPS)},
year={2024}
}

@inproceedings{blip2,
  title={BLIP-2: Bootstrapping Language-Image Pre-training with Frozen Image Encoders and Large Language Models},
  author={Junnan Li and Dongxu Li and Silvio Savarese and Steven C. H. Hoi},
  booktitle={International Conference on Machine Learning},
  year={2023},
  url={https://api.semanticscholar.org/CorpusID:256390509}
}

@article{dai2022toyotasmarthomeuntrimmed,
  title        = {Toyota Smarthome Untrimmed: Real-World Untrimmed Videos for Activity Detection},
  author       = {Dai, Rui and Das, Srijan and Sharma, Saurav and Minciullo, Luca and Garattoni, Lorenzo and Bremond, Francois and Francesca, Gianpiero},
  journal      = {IEEE Transactions on Pattern Analysis and Machine Intelligence},
  year         = {2022},
  doi          = {10.1109/TPAMI.2022.3169976}
}

@inproceedings{zhang2025exo2egovlm,
      title={Exo2Ego: Exocentric Knowledge Guided MLLM for Egocentric Video Understanding}, 
      author={Haoyu Zhang and Qiaohui Chu and Meng Liu and Haoxiang Shi and Yaowei Wang and Liqiang Nie},
      year={2026},
      booktitle = {Proceedings of the AAAI Conference on Artificial Intelligence}
}

@inproceedings{abnar2020attentionrollout,
      title={Quantifying Attention Flow in Transformers},
      author={Abnar, Samira and Zuidema, Willem},
      booktitle={Annual Meeting of the Association for Computational Linguistics},
      year={2020}
}

@book{kldivergence_information_theory,
author = {Cover, Thomas M. and Thomas, Joy A.},
title = {Elements of Information Theory (Wiley Series in Telecommunications and Signal Processing)},
year = {2006},
isbn = {0471241954},
publisher = {Wiley-Interscience},
address = {USA}
}

@article{miech2019howto100m,
  title   = {HowTo100M: Learning a Text-Video Embedding by Watching Hundred Million Narrated Video Clips},
  author  = {Miech, Antoine and Zhukov, Dimitri and Alayrac, Jean-Baptiste and Tapaswi, Makarand and Laptev, Ivan and Sivic, Josef},
  journal = {arXiv preprint arXiv:1906.03327},
  year    = {2019}
}

@article{tvbench,
      title={TVBench: Redesigning Video-Language Evaluation}, 
      author={Daniel Cores and Michael Dorkenwald and Manuel Mucientes and Cees G. M. Snoek and Yuki M. Asano},
      year={2024},
      journal={arXiv preprint arXiv:2410.07752}
}

@article{temporalbench,
      title={TemporalBench: Benchmarking Fine-grained Temporal Understanding for Multimodal Video Models}, 
      author={Mu Cai and Reuben Tan and Jianrui Zhang and Bocheng Zou and Kai Zhang and Feng Yao and Fangrui Zhu and Jing Gu and Yiwu Zhong and Yuzhang Shang and Yao Dou and Jaden Park and Jianfeng Gao and Yong Jae Lee and Jianwei Yang},
      year={2024},
      journal={arXiv Preprint arXiv:2410.10818}
}

@misc{motionbench2025,
      title={MotionBench: Benchmarking and Improving Fine-grained Video Motion Understanding for Vision Language Models}, 
      author={Wenyi Hong and Yean Cheng and Zhuoyi Yang and Weihan Wang and Lefan Wang and Xiaotao Gu and Shiyu Huang and Yuxiao Dong and Jie Tang},
      booktitle={Proceedings of the IEEE/CVF Conference on Computer Vision and Pattern Recognition (CVPR)},
      year={2025}
}

@inproceedings{zohar2025apollo,
      title={Apollo: An Exploration of Video Understanding in Large Multimodal Models},
      author={Zohar, Orr and Wang, Xiaohan and Dubois, Yann and Mehta, Nikhil and Xiao, Tong and Hansen-Estruch, Philippe and Yu, Licheng and Wang, Xiaofang and Juefei-Xu, Felix and Zhang, Ning and Yeung-Levy, Serena and Xia, Xide},
      booktitle={Proceedings of the IEEE/CVF Conference on Computer Vision and Pattern Recognition},
      year={2025}
}
}

\clearpage
\maketitlesupplementary

\appendix
\section{Appendix}
In Section \ref{sec:supp_egoperceptiondetails}, we detail the data curation pipeline of \benchname~benchmark. In Section \ref{sec:supp_egoperceptionexamples}, we present example video-MCQ pairs from \benchname~across all categories. In Section \ref{sec:supp_studentinstructiondetails}, we detail how we leverage the ego-observing teacher VLM to generate instructions that supervise the exo-observing student VLM.

\begin{figure}[h]
    \centering

    \includegraphics[width=\linewidth]{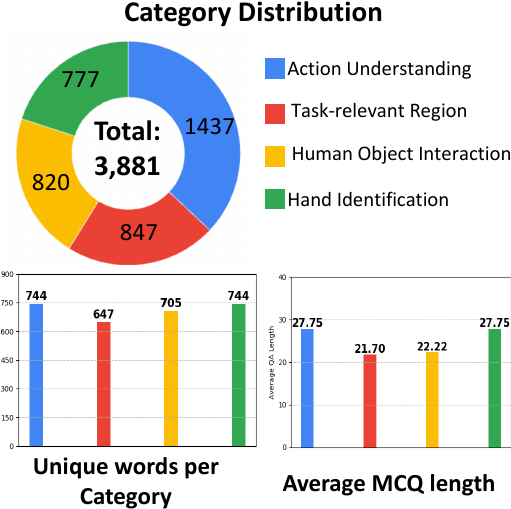}
    \caption{\textbf{\benchname Statistics.} We present the count of MCQs in each category, along with the per-category unique word count and average MCQ length.}
    \label{fig:supp_egoperception_stats}
\end{figure}

\subsection{\benchname~Curation Details}
\label{sec:supp_egoperceptiondetails}
Detailed statistics of \benchname~can be found in Figure \ref{fig:supp_egoperception_stats}. In the following, we provide additional details on the construction of our \benchname~benchmark.

\noindent\textbf{Scene object parsing.}\quad The atomic action annotations from EgoExo4D provide dense details about actions and HOIs in each video, but they lack scene content. Extracting scene objects enables the generation of more diverse question types, and facilitates the creation of more challenging hard negative answers. To obtain scene objects for a given keystep clip, we use CogAgent~\cite{hong2023cogagent} with a sliding window over the ego viewpoint, generating image captions at 5fps. The prompt we use for CogAgent is ``Describe the scene and what objects are visible in the scene". We then use an LLM to parse the captions into a list of scene objects, which we use as input alongside atomic action annotations, to GPT-4o for generating MCQs.

\noindent\textbf{MCQ Generation.}\quad
Our benchmark comprises \textit{four} distinct categories of multiple-choice questions, manually designed to evaluate egocentric understanding. For each video clip, we leverage GPT-4o~\cite{openai2024gpt4ocard} with category-specific prompts to generate a single question per category. The four categories are defined as: \textbf{Action Understanding (Action)}, which assesses comprehension of the actions being performed; \textbf{Task-relevant Region (Task-R)}, which evaluates understanding of spatial areas where the primary keystep action is being performed; \textbf{Human Object Interactions (HOI)}, which measures understanding of human and object interactions; and \textbf{Hand Identification (Hand)}, which evaluates ability to distinguish the specific hand used to perform an action.

\begin{table}[h]
    \renewcommand{\arraystretch}{1.2}
    \centering
    \caption{\textbf{Human verification criteria.} Annotators rate each video–MCQ pair based on three criteria: question correctness, exocentric visibility, and the presence of at least one hard negative answer. Final scores range from 1-3 depending on how many criteria are satisfied.}
    \resizebox{0.98\linewidth}{!}{
    \begin{tabular}{>{\centering\arraybackslash}p{2.5cm}|p{5.5cm}}
        \hline
        \textbf{Criterion} & \multirow{1}{*}{\parbox{\linewidth}{\centering \textbf{Given Definition}}} \\
        \hline
        \multirow{4}{*}{Correctness} &
        The video-MCQ matches the ground-truth atomic actions, the action truly occurs in the video, and the question is coherent and non-hallucinated. \\
        \hline

        Exocentric visibility &
        The referenced action is observable from the exocentric viewpoint; occlusions or camera placement do not hide the key context. \\
        \hline
        \multirow{3}{*}{Hard negatives} &
        The answer set includes at least one plausible but incorrect alternative that requires viewing the video to eliminate. \\
        \hline
        \rowcolor{LightBlue}
        \multirow{2}{*}{\textbf{Scoring}} &
        \textbf{3 = all criteria satisfied; 2 = any two satisfied; 1 = any one satisfied.} \\
        \hline
    \end{tabular}}
    \label{tab:human_verification_criteria}
\end{table}

\begin{figure*}[t!]
    \centering
    \includegraphics[width=\linewidth]{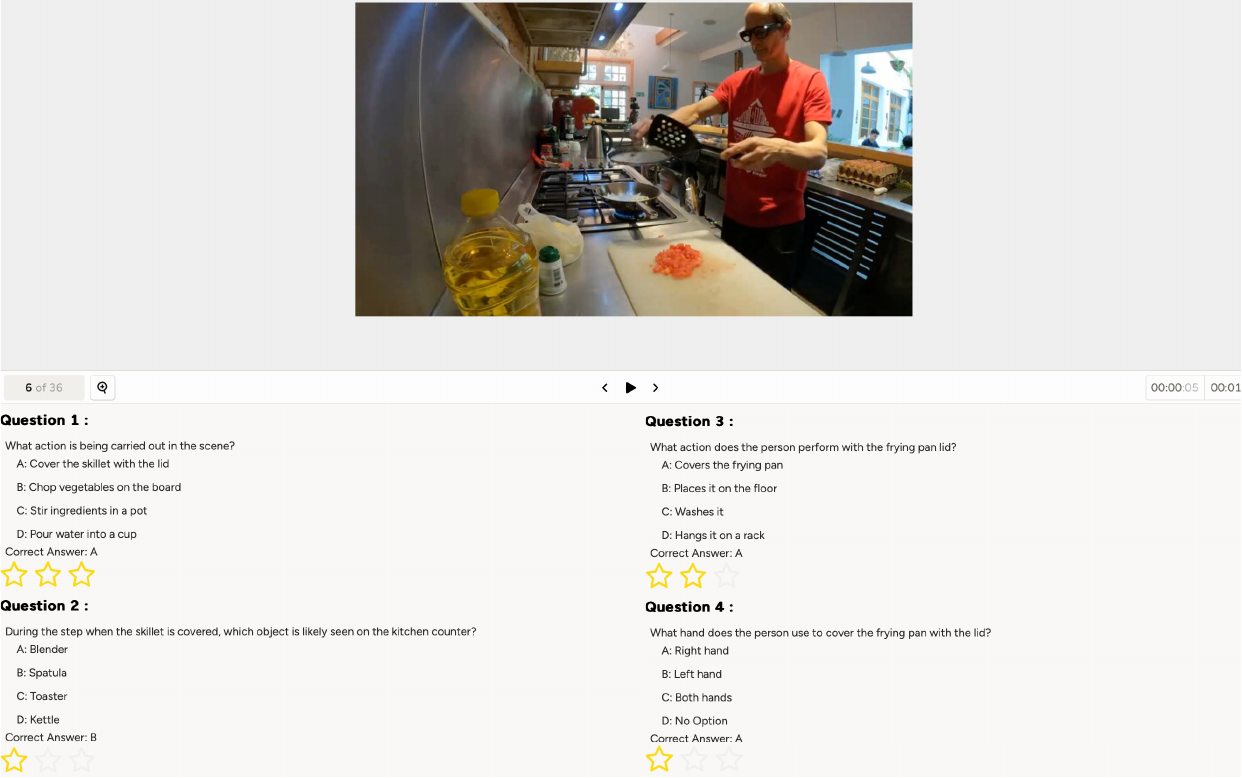}
    \caption{\textbf{Interface used for quality verification.} Human verifiers were asked to rate each generated video-MCQ pair on a scale of 1-3 (the stars in the interface correspond to the selected score).}
    \label{fig:supp_humanverification_interface}

\end{figure*}

\noindent\textbf{Human quality verification}\quad Previous research~\cite{tvbench, temporalbench, motionbench2025} has identified common issues with LLM-generated MCQs, including hallucinated questions and easy negative answers. Our benchmark presents an additional unique challenge: as questions are generated using annotations derived from the ego view, they may not always be answerable from the exo perspective due to occlusions or camera placement. To address these concerns, we first filter out generated MCQs that the LLM itself was not confident in, resulting in 5,689 remaining video-MCQ pairs. We then employ four human annotators to manually verify the remaining video-MCQ pairs based on three criteria: question correctness, exocentric visibility, and presence of hard negative answers. Each video-MCQ pair is scored from 1-3, corresponding to the number of criteria met. This scoring protocol, as well as definitions of each criteria given to annotators, is illustrated in Table \ref{tab:human_verification_criteria}. We retain only those video-MCQ pairs with a score of 3, resulting in a total of 3,881 MCQs used in our benchmark. All scoring was done through an interface built using Label Studio, which we present in Figure \ref{fig:supp_humanverification_interface}.

\subsection{\benchname~Examples}
\label{sec:supp_egoperceptionexamples}
\begin{figure*}[t]
    \centering
    \includegraphics[width=0.88\linewidth]{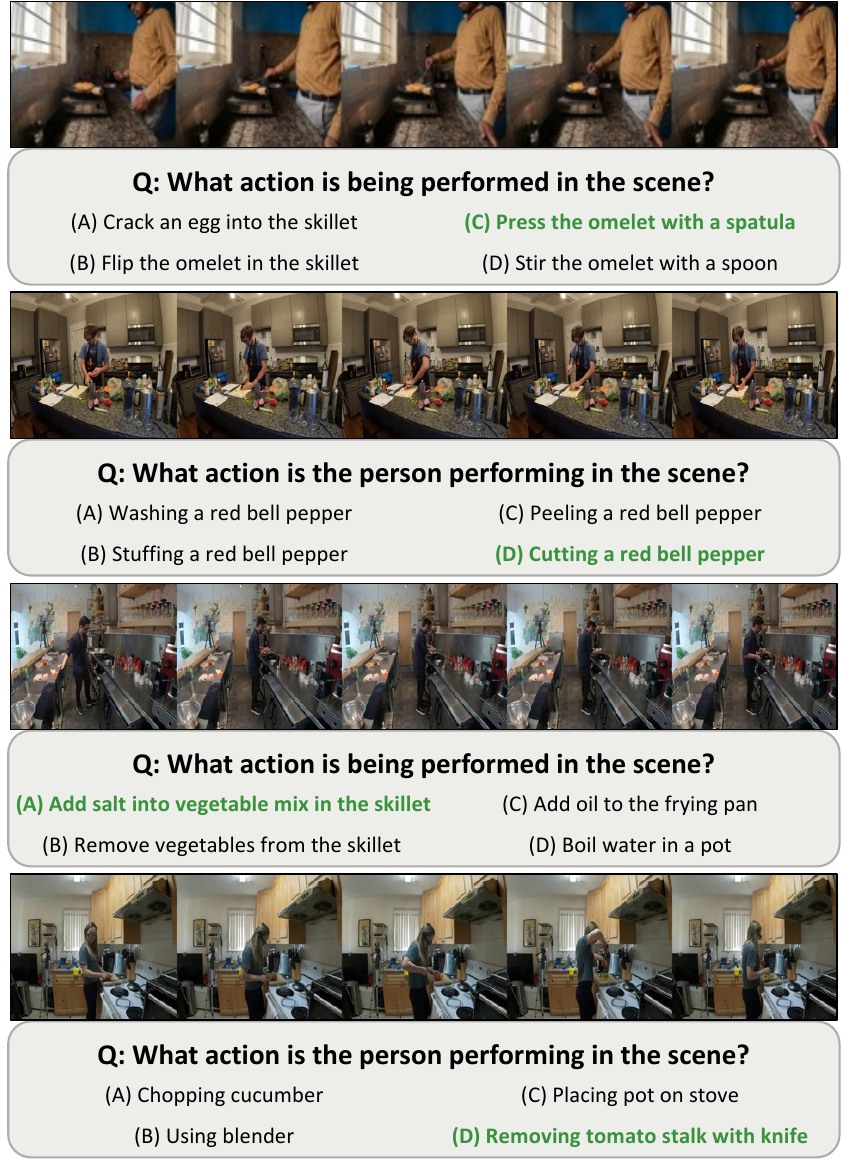}
    \caption{\textbf{Examples from the Action Understanding category of \benchname.} The correct answer is shown in green.}
    \label{fig:supp_bench_actionund}
\end{figure*}

\begin{figure*}[t]
    \centering
    \includegraphics[width=0.88\linewidth]{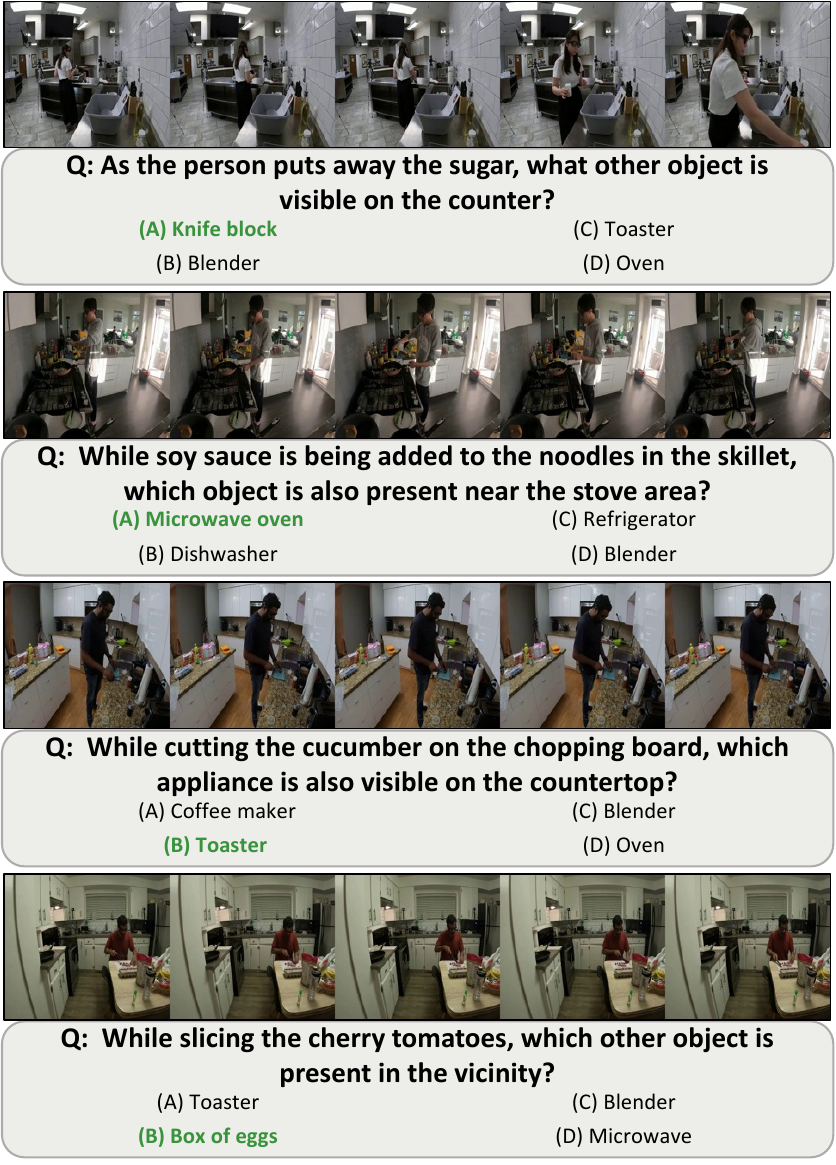}
    \caption{\textbf{Examples from the Task-relevant Region category of \benchname.} The correct answer is shown in green.}
    \label{fig:supp_bench_taskregion}
\end{figure*}

\begin{figure*}[t]
    \centering
    \includegraphics[width=0.88\linewidth]{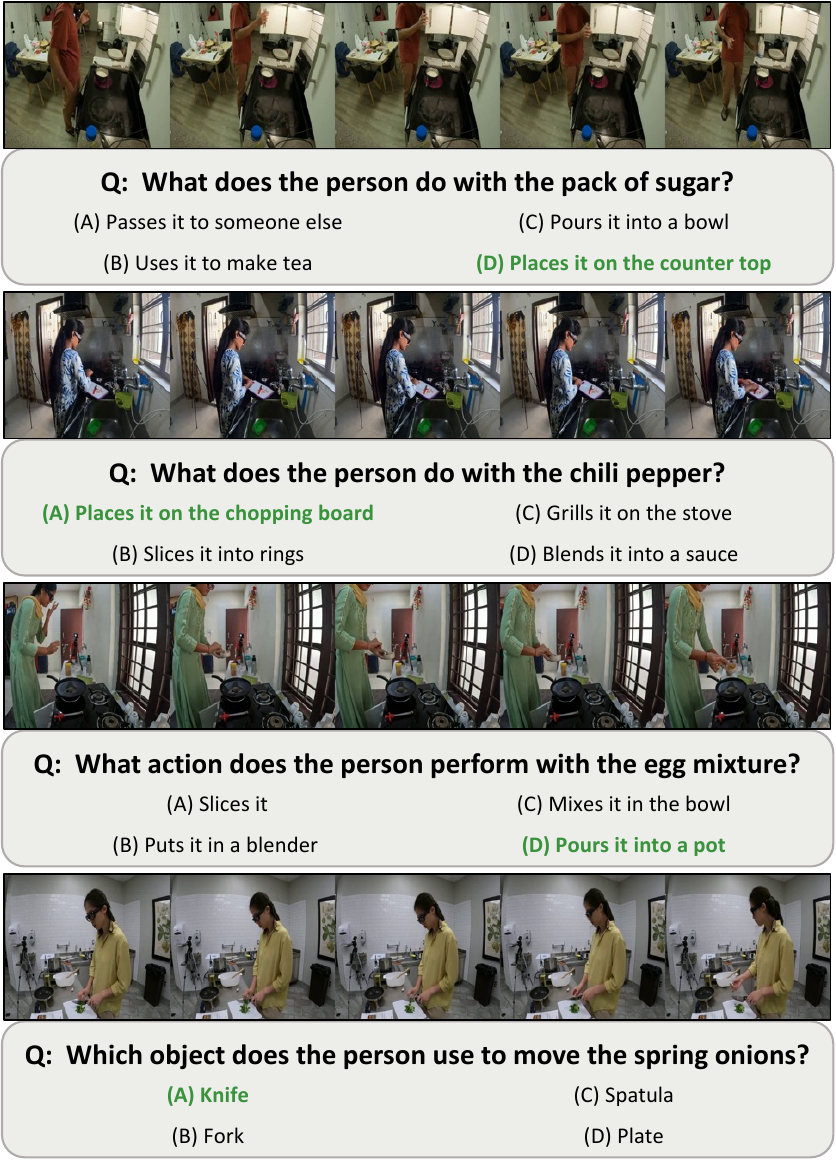}
    \caption{\textbf{Examples from the Human Object Interaction category of \benchname.} The correct answer is shown in green.}
    \label{fig:supp_bench_hoi}
\end{figure*}

\begin{figure*}[t]
    \centering
    \includegraphics[width=0.88\linewidth]{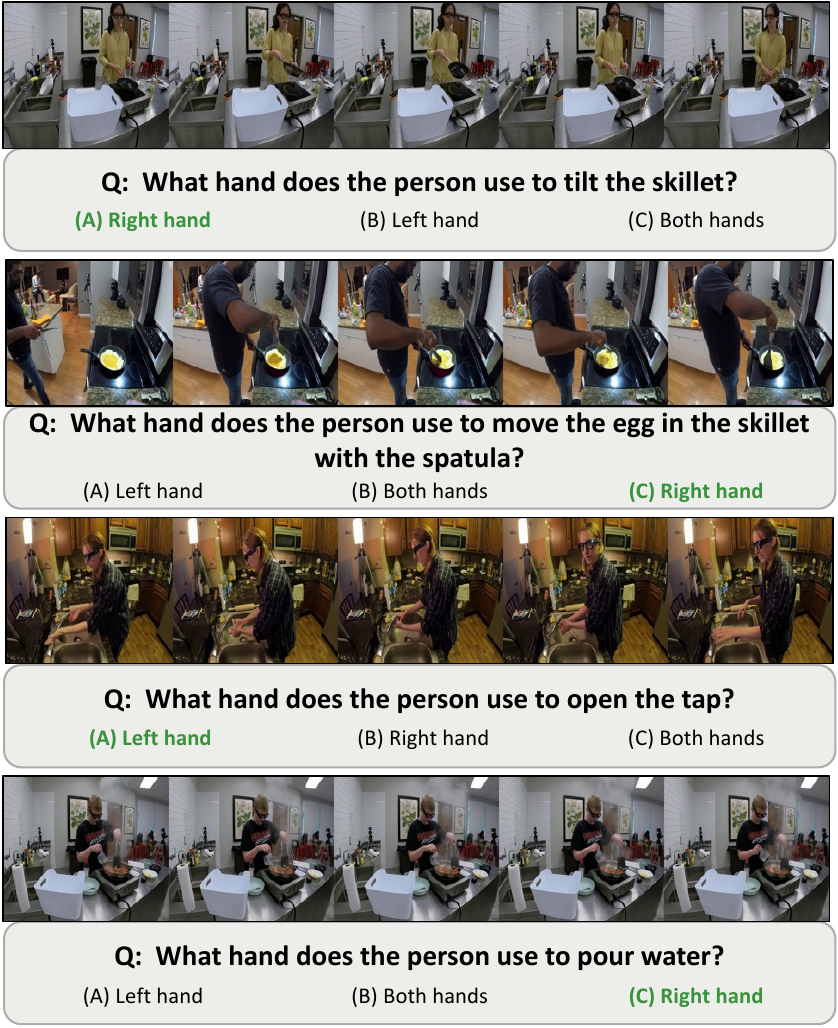}
    \caption{\textbf{Examples from the Hand Identification category of \benchname.} The correct answer is shown in green.}
    \label{fig:supp_bench_handident}
\end{figure*}

In this section, we present example video-MCQ pairs from \benchname~for each of the four categories in the benchmark. Examples from the Action Understanding category are presented in Figure \ref{fig:supp_bench_actionund}, examples from the Task-relevant Region category are presented in Figure \ref{fig:supp_bench_taskregion}, examples from the Human Object Interaction category are presented in Figure \ref{fig:supp_bench_hoi}, and examples from the Hand Identification category are presented in Figure \ref{fig:supp_bench_handident}.

\subsection{Teacher Instruction Generation for Student}
\label{sec:supp_studentinstructiondetails}
We obtain training instructions to supervise the student VLM using the teacher model and 9k ego–exo video pairs from the keystep recognition training set of EgoExo4D~\cite{cvpr2025egoexo4d}. To generate these training instructions, we obtain three natural language queries for each of the 9k egocentric videos, each query is sampled from each of three categories: descriptive, temporal, and spatial. These categories are designed to probe complementary aspects of video understanding so that the teacher VLM produces answers that reflect different dimensions of the underlying activity. For each of these sampled queries, the \emph{egocentric video} and the query are passed to the frozen teacher VLM, which produces an answer. Then, this query and answer is paired with the corresponding \emph{exocentric video} to be used as a training instruction to supervise the student VLM. Below we list the full sets of prompts used in each category.

\noindent\textbf{Descriptive Queries.}\quad
Descriptive queries encourage the teacher to summarize the foreground actions, interactions, and overall activity context in the video. A descriptive query focuses on the core action semantics of the scene rather than temporal structure or object arrangement.

\begin{itemize}
\item Describe what is happening in this video in detail.
\item What events are taking place in this video in detail?
\item Describe the main actions shown in the video in detail.
\item What activity is being performed in the video in detail?
\item Provide a detailed description of what occurs in this clip.
\end{itemize}

\noindent\textbf{Temporal Queries}\quad
Temporal queries target the sequence structure of the activity. These questions direct the teacher to capture transitions, changes, and the overall flow from start to finish. Unlike descriptive queries, the emphasis here is on temporal understanding rather than pure descriptive tasks.

\begin{itemize}
\item Summarize the sequence of events shown in this video.
\item Outline the order of events shown in this video.
\item Summarize the progression of events from beginning to end.
\end{itemize}

\noindent\textbf{Spatial Queries}\quad
Spatial queries focus on the objects, tools, and physical layout present in the scene. These questions force the teacher to identify and describe the objects and interactions involved in the activity, and how they are arranged or used. The resulting answers highlight object presence and their use.

\begin{itemize}
\item What are the primary objects in the scene relevant to the action being performed?
\item Where are the main objects and people located in the scene?
\item Describe the objects in the video and how they are used.
\item What objects are visible throughout the video?
\end{itemize}

\end{document}